\documentclass{article}

\usepackage{microtype}
\usepackage{graphicx}
\usepackage{subcaption}
\usepackage{booktabs} % for professional tables
\usepackage{enumitem}
\usepackage{tabularx,array}
\newcolumntype{Y}{>{\centering \arraybackslash}p{0.17\linewidth}} % 你可以调 0.20~0.28
\usepackage{hyperref}
\usepackage{acro}
\usepackage{caption}
\usepackage[percent]{overpic}
\usepackage{multirow}
\usepackage[accepted]{icml2026}

\usepackage{amsmath}
\usepackage{amssymb}
\usepackage{mathtools}
\usepackage{amsthm}
\usepackage{wrapfig}
\usepackage[normalem]{ulem}
\useunder{\uline}{\ul}{}

\usepackage[capitalize,noabbrev]{cleveref}

\theoremstyle{plain}

\theoremstyle{definition}

\theoremstyle{remark}

\usepackage[textsize=tiny]{todonotes}
\usepackage[table]{xcolor}

\definecolor{rankone}{rgb}{0.65, 0.89, 0.94}    
\definecolor{ranktwo}{rgb}{0.85, 0.91, 0.95}    
\definecolor{rankthree}{rgb}{0.96, 0.96, 0.98}  

\definecolor{rankblue}{RGB}{52, 84, 132}

\definecolor{ranklightblue}{RGB}{210, 225, 245}
\icmltitlerunning{Automatic Layer Selection for Hallucination Detection}
\DeclareAcronym{saplma}{short = SAPLMA, long = Statement Accuracy Prediction based on Language Model Activations}
\DeclareAcronym{lid}{short = LID, long = Local Intrinsic Dimensionality}
\DeclareAcronym{eigenscore}{
  short = EigenScore,
  long  = EigenScore,
  first-style = short
}
\DeclareAcronym{our_method}{
  short = FEPoID,
  long  = First Effective Peak of Intrinsic Dimension
}
\DeclareAcronym{llama}{
  long = LlaMA-3.1-8B-Instruct,
  short  = LlaMA-Instruct
}
\DeclareAcronym{mistral}{
  long =  Mistral-7B-Instruct-v0.3,
  short  =  Mistral-Instruct
}

\begin{document}

\twocolumn[
  \icmltitle{Automatic Layer Selection for Hallucination Detection}



  \icmlsetsymbol{equal}{*}

  \begin{icmlauthorlist}
    \icmlauthor{Xinpeng Wang}{uva}
    \icmlauthor{William X. Cao}{comp}
    \icmlauthor{Andrew Gordon Wilson}{nyu}
    \icmlauthor{Zhe Zeng}{uva}
  \end{icmlauthorlist}

  \icmlaffiliation{uva}{University of Virginia}
  \icmlaffiliation{comp}{Walrus Security}
  \icmlaffiliation{nyu}{New York University}

  \icmlcorrespondingauthor{Xinpeng Wang}{hqr4gx@virginia.edu}

  \icmlkeywords{Machine Learning, ICML}

  \vskip 0.3in
]



\printAffiliationsAndNotice{WC contributed to this work as an independent researcher.}  % no special notice (required even if empty)

\begin{abstract}
Recent studies on hallucination detection have shown that hallucination-related signals are more strongly encoded in intermediate layers than in the final layer of large language models~(LLMs). Although a growing body of work has sought to exploit this property for hallucination detection, how to automate the selection of high-performing layers remains underexplored, and principled methods for this purpose are still lacking. To address this gap, we first propose several hypotheses for why such signals emerge in intermediate layers and evaluate corresponding criteria for automatic layer selection across diverse LLM architectures, scales, and tasks, covering both question answering and summarization hallucination detection benchmarks. However, we find that none of these criteria consistently delivers satisfactory performance. We therefore propose a new selection criterion, \ac{our_method}, which consistently identify optimal or near-optimal layers and outperforms both the aforementioned criteria and existing hallucination detection baselines. \ac{our_method} is training-free and incurs negligible computational overhead. In addition, we study the generation behaviors of LLMs and introduce a simple yet effective truncation strategy,  which further amplifies hallucination-related signals and substantially improves overall detection performance. Code is publicly available at \href{https://github.com/DesoloYw/Automatic-Layer-Selection-for-Hallucination-Detection.git}{https://github.com/DesoloYw/Automatic-Layer-Selection-for-Hallucination-Detection.git}

\end{abstract}
% \input{sections/introduction.tex}
% \input{sections/related_work}
% \input{sections/motivation.tex}
% % \input{sections/preliminary.tex}
% \input{sections/method.tex}
% \input{sections/experiment.tex}
% \input{main_body.tex}

% \zz{maybe we can change the title to be more specific for hallucination detection}
\section{Introduction}

\begin{figure}[t]
\centering

% subcaption font (apply to BOTH keys)
\captionsetup[sub]{font=scriptsize, labelfont=scriptsize}
\captionsetup[subfigure]{font=scriptsize, labelfont=scriptsize}

\resizebox{\linewidth}{!}{%
  \begin{minipage}{\linewidth}
    \centering
    \begin{tabular}{@{}c@{}c@{}c@{}c@{}}
      \raisebox{0pt}[0pt][0pt]{\includegraphics[height=3.0cm]{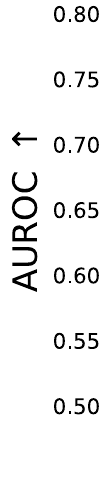}} &
      \subcaptionbox{LLaMA-3.1-8B-Instruct\label{fig:panel2}}{%
        \includegraphics[height=3.0cm,keepaspectratio]{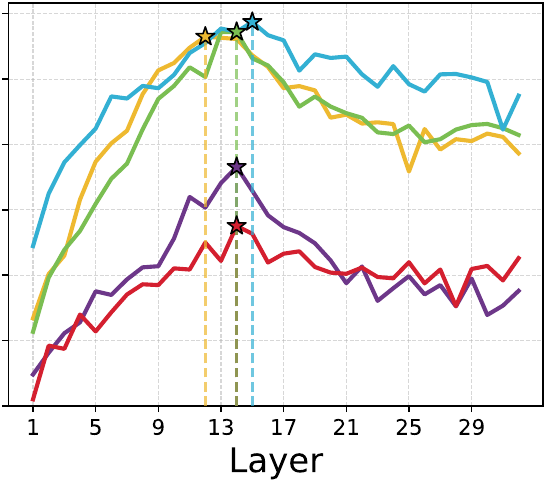}} &
      \raisebox{0pt}[0pt][0pt]{\includegraphics[height=3.0cm]{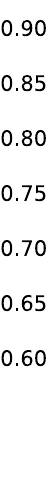}} &
      \subcaptionbox{Mistral-7B-Instruct-v0.3\label{fig:panel4}}{%
        \includegraphics[height=3.0cm,keepaspectratio]{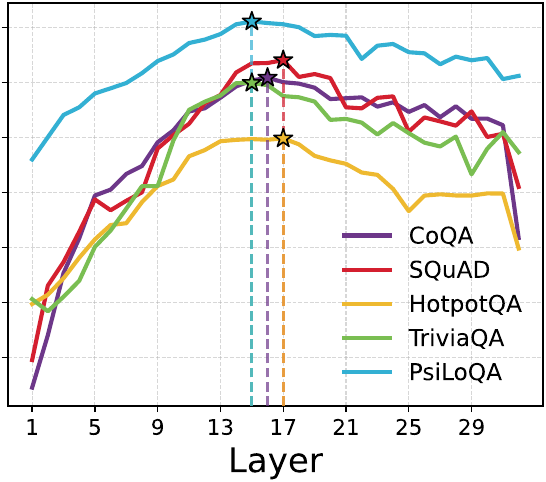}}
    \end{tabular}

    \vspace{2pt}

    \subcaptionbox{Performance boost with our layer-selection based strategies\label{fig:panelbar}}{%
      \includegraphics[width=0.9\linewidth,keepaspectratio]{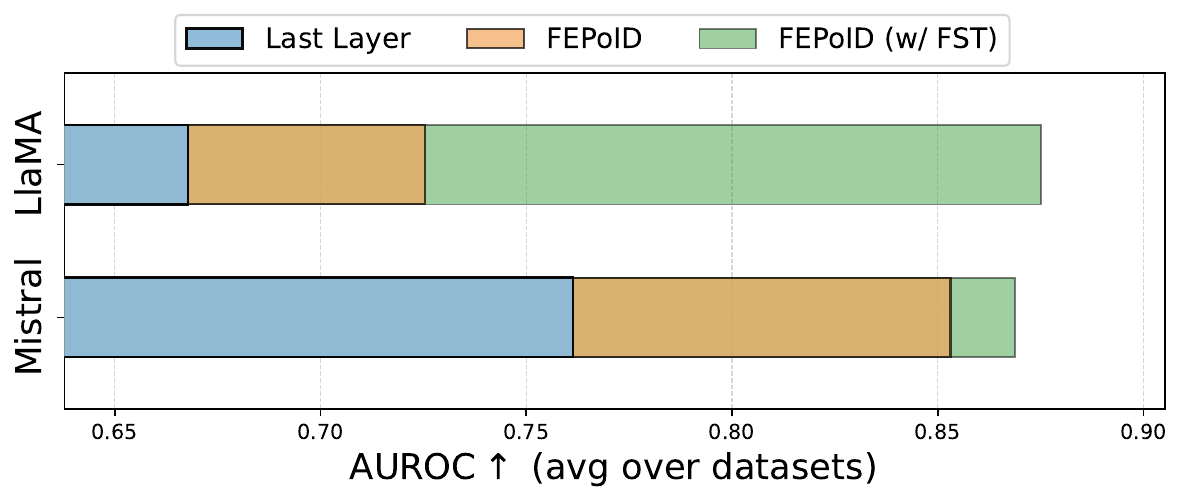}}
  \end{minipage}%
}

\caption{Hallucination detection performance under a unified experimental setting.
For all experiments, we extract last-token representations from each layer and train an MLP classifier for hallucination detection. \textbf{(TOP):} Layer-wise AUROC under oracle training, where the best-performing layer (starred) consistently lies in the intermediate layers. \textbf{(Bottom):} Mean AUROC averaged across datasets under different layer-selection strategies.
FEPoID consistently outperforms the last-layer heuristic, and its combination with the truncation strategy yields further gains across models.}
\label{fig:motivation}
\end{figure}

Hallucination detection is a critical challenge for deploying large language models (LLMs) in real-world applications, as LLMs can produce fluent yet factually incorrect or internally inconsistent outputs. Detecting such hallucinations without modifying or fine-tuning the underlying model is therefore an important practical problem~\citep{huang2025survey, semantic_entropy_nature, li-etal-2024-dawn}. Prior work has approached hallucination detection using uncertainty-based estimates~\citep{semantic_entropy_nature,predictive_entropy_and_ln_entropy} or verbalized uncertainty~\citep{xiong2024can,zhou-etal-2023-navigating}. In contrast, recent studies show that hallucination-related signals are more strongly encoded in the internal representations of LLMs than in their final outputs, motivating the use of hidden states for hallucination detection~\citep{llmknows,saplma,eigenscore,LID,ji-2024-llm}.

However, most existing approaches either select a predetermined intermediate layer in a data- and task-agnostic manner, or fully evaluate each candidate layer, which is impractical due to the computational cost.
% for real-world applications. 
To demonstrate why layer selection is challenging, 
we visualize in \cref{fig:motivation} the best-performing layer for hallucination detection, where it consistently lies in the intermediate layers, but its exact location varies substantially across datasets and model architectures. This variability 
% makes fixed heuristics impractical and 
motivates the central question of this paper: \emph{can we design a practical and principled criterion that automatically identifies the most informative intermediate layer for hallucination detection?}

Throughout this work, we study this problem within the hidden-state probing framework, where the pretrained LLM is kept frozen and a lightweight multi-layer perceptron~(MLP) is trained on representations extracted from a selected layer for hallucination detection. 
% To find practical layer-selection criterion, 
We first propose several hypotheses for why hallucination-related signals emerge in intermediate layers. 
Guided by these hypotheses,
we systematically evaluate a diverse set of candidate layer-selection criteria including information-theoretic, gradient-based and geometric criteria across diverse LLM architectures, scales, and tasks, covering both question answering and summarization hallucination detection benchmarks.
The empirical results show that, none of these criteria can efficiently and reliably identify high-performing layers.
% We find that only validation loss yields consistently strong performance; however, it requires MLP training on each layer, making it computationally expensive and impractical.

% Instead, we propose a new selection criterion, \acf{our_method}, motivated by our analysis of the intrinsic-dimension evolution cross layers. Specifically, we observe a shared pattern across various settings where an early peak in the intermediate layers followed by a later, often higher peak near the output. 
% While higher intrinsic dimension generally indicates richer representations, the early peak is likely associated with abstract semantic information that is most relevant to hallucination detection, whereas the later peak primarily captures surface-level complexity making it less informative for this task.
Instead, inspired by our empirical observation of how intrinsic dimension evolves across layers, we propose a new selection criterion: the \underline{F}irst \underline{E}ffective \underline{P}eak \underline{o}f \underline{I}ntrinsic \underline{D}imension (FEPoID).
Across models and datasets, we observe a recurring pattern in which the intrinsic dimension first peaks in the intermediate layers and later reaches another, often higher, peak near the output layers.
We hypothesize that these two peaks reflect different forms of representational complexity:
the earlier peak captures abstract semantic information that is especially relevant to hallucination detection, whereas the later peak primarily captures surface-level complexity making it less informative for this task.
This is further supported by our empirical results: selecting the first effective ID peak consistently identifies optimal or near-optimal layers, enabling FEPoID to outperform the aforementioned criteria and achieve stronger hallucination detection performance than existing baselines.

While choosing an appropriate layer is crucial for effective hidden-state probing, performance in hallucination detection also critically depends on the \emph{token position} used for representation extraction. A common heuristic is to extract representations at the last generated token, motivated by the autoregressive property that this token can attend to the entire context. However, recent studies have shown that last-token representations are sensitive to noise introduced near the end of the generated sequence~\citep{springer2025repetition,lee2025nvembed} and often underperform on downstream tasks~\citep{llmknows}. 
This raises the second research question of our work: \emph{can we identify a simple, supervision-free rule that yields informative representations?}

% In this study, we further 
We explore this question by evaluating extracted representations at the last token of the first generated sentence, identified using a simple, rule-based First-Sentence Truncation (FST). Through extensive experiments, we find that representations extracted at the end of the first sentence consistently yield stronger detection performance as shown in \cref{fig:motivation}. 
% Meanwhile \wc{decide if this is the right word}, 
This is motivated by our observation that the representations at the last generated tokens are often degraded by end-of-sequence noise arising from degenerate repetition, inconsistent continuation, and semantic drift. In addition, FST consistently improves the performance of various hallucination detection baselines, indicating that its effectiveness is not tied to any specific modeling assumptions, but instead comes from systematically reducing noise introduced during late-stage generation.

In summary, our main contributions are threefold:
\begin{enumerate}[label=\roman*),noitemsep,topsep=0pt]
    \item  We provide the first systematic evaluation of criteria that have been shown in prior work to correlate with downstream performance, as well as criteria used for layer-selective fine-tuning, both of which remain underexplored for practical layer selection.
    \item We introduce \ac{our_method}, a simple and efficient criterion that automatically selects near-optimal intermediate layers across various datasets and pretrained models. 
    \item We revisit token-position choices for hidden-state probing and show that extracting representations at the last token of the first generated sentence consistently outperforms the common last-token heuristic. Moreover, applying FST improves all hallucination detection baselines considered in this work, indicating a method-agnostic benefit from mitigating noise introduced during late-stage generation.

\end{enumerate}

% \begin{table}[t]
% \centering
% \caption{Hypotheses and corresponding criteria. 
% % for intermediate-layer selection.
% }
% \label{tab:hypotheses}
% \small
% \renewcommand{\arraystretch}{1.4}  
% \scalebox{1}{
% \begin{tabular}{lll}
% \toprule
% \multicolumn{2}{l}{\textbf{Hypothesis}}     & \textbf{Criteria}                                                       \\ \midrule
% (i)   & Rich semantic information          & RankMe                                                                  \\
% (ii)  & Task-aligned features & \begin{tabular}[l]{@{}l@{}}Validation Loss, \\RGN, SNR\end{tabular} \\
% (iii) & Information compression                         & Curvature                                                               \\
% (iv)  & High effective information capacity    & ID, FEPoID                                                       \\ \bottomrule
% \end{tabular}
% }
% \vskip -0.1in
% \end{table}

\section{Related Work}
\label{sec:related_work}
% \paragraph{Why are Intermediate Layers Better}
% \citet{layer_by_layer} shows that intermediate layers often outperform final layers across language and vision, and evaluates curvature- and rank-based metrics. However, their work primarily focuses on using these metrics to analyze why intermediate layers can be more effective, rather than addressing the layer-selection problem for downstream tasks, which is the focus of our study.

\paragraph{Criteria for Intermediate Layer Selection}
\citet{layer_by_layer} show that intermediate layers can encode rich information across various architectures and domains.
\citet{surgical_finetuning} study selective fine-tuning under distribution shifts and use supervised signals such as the relative gradient norm (RGN) and signal-to-noise ratio (SNR) to choose which layers to update; 
we repurpose these criteria for selecting probe layers and show that they perform poorly in our setting. \citet{curvature} introduce curvature to quantify layer-wise flattening of sentence embeddings, which has been shown to correlate with downstream performance~\citep{layer_by_layer}.
\citet{earthID} show that the ID of learned geographic representations is positively correlated with downstream task performance, and can capture meaningful structural properties of the data. \citet{cheng2025emergence} show that layers near the maximum ID tend to be the first to transfer effectively to downstream tasks.
However, these criteria have not been systematically studied for automatic layer selection and for hallucination detection, which is the focus of this work.

\begin{table}[t]
\centering
\caption{Hypotheses and corresponding criteria. 
% for intermediate-layer selection.
}
\label{tab:hypotheses}
\small
\renewcommand{\arraystretch}{1.4}  
\scalebox{1}{
\begin{tabular}{lll}
\toprule
\multicolumn{2}{l}{\textbf{Hypothesis}}     & \textbf{Criteria}                                                       \\ \midrule
(i)   & Rich semantic information          & RankMe                                                                  \\
(ii)  & Task-aligned features & \begin{tabular}[l]{@{}l@{}}Validation Loss, \\RGN, SNR\end{tabular} \\
(iii) & Information compression                         & Curvature                                                               \\
(iv)  & High effective information capacity    & ID, FEPoID                                                       \\ \bottomrule
\end{tabular}
}
\vskip -0.1in
\end{table}

\paragraph{Intermediate Layer Trials}
Existing works that leverage pretrained model representations for downstream tasks commonly  select a fixed intermediate (often the middle) layer~\citep{eigenscore}, or restrict evaluation to a predefined subset of layers (e.g., middle layer and final layer)~\citep{liu2024uncertainty,ahdritz2024distinguishing,ji-2024-llm}. Some prior works explore layer selection by evaluating a broader but still limited set of candidates. For example, \citet{saplma} probe a small grid of layers (e.g., layer 16, 20, 24, 28 and the last layer in a 32-layer model) and observe that certain intermediate layers perform best. Similarly, \citet{llmknows} evaluate hallucination detection across a sparse set of layers (layer 1, 6, 11, \dots, 31) and find that layers in the middle range tend to be more informative. 
While these approaches can outperform the last layer alone, they also highlight the lack of a practical and principled way to reliably identify strong intermediate layers.

% \vspace{-1em}
\paragraph{Token Position for Extraction}
A widely used heuristic is to extract the hidden state of the \emph{last generated token}, motivated by the autoregressive property that it attends to all preceding context, but this representation is sensitive to end-of-sequence noise~\citep{springer2025repetition,lee2025nvembed}.
An alternative extracts features from the last prompt token; however, the unidirectional nature of decoder-only LLMs prevents this representation from reflecting correctness differences across sampled outputs~\citep{slobodkin2023the}. Another option is to average representations across token positions, but prior studies~\citep{li2025making,zhang-etal-2025-language} show that under causal attention, mean pooling is less effective than last-token features, since earlier tokens cannot attend to future tokens.

\citet{llmknows} evaluate multiple token positions---including the last generated token, the last prompt token, and the ``exact answer'' token---and found that representations at the exact-answer position achieve the best downstream performance within the hidden-state probing framework. This suggests that the truthfulness-related information is more concentrated at answer-aligned token positions. However, locating the ``exact answer'' token requires the ground-truth answer as a reference, which is impractical since it is a chicken-and-egg problem. Also, for open-ended scenarios, identifying the ``exact answer'' token typically relies on auxiliary LLMs, incurring additional computational overhead.

\section{Layer-Selection Criteria}
\label{sec:method}

% While any intermediate layer of a pretrained model can potentially yield strong downstream performance, it remains unclear which layers are most suitable in practice.
% Our goal is therefore to identify principled criteria that can distinguish such layers without relying on exhaustive downstream training.
% preamble:
% \usepackage{tabularx}

% \begin{table}[t]
% \centering
% \caption{Hypotheses and corresponding criteria. 
% % for intermediate-layer selection.
% }
% \label{tab:hypotheses}
% \small
% \renewcommand{\arraystretch}{1.4}  
% \scalebox{1}{
% \begin{tabular}{lll}
% \toprule
% \multicolumn{2}{l}{\textbf{Hypothesis}}     & \textbf{Criteria}                                                       \\ \midrule
% (i)   & Rich semantic information          & RankMe                                                                  \\
% (ii)  & Task-aligned features & \begin{tabular}[l]{@{}l@{}}Validation Loss, \\RGN, SNR\end{tabular} \\
% (iii) & Information compression                         & Curvature                                                               \\
% (iv)  & High effective information capacity    & ID, FEPoID                                                       \\ \bottomrule
% \end{tabular}
% }
% \vskip -0.1in
% \end{table}
We seek criteria that capture the relationship between intermediate-layer representations and downstream task performance.
% Our choices are guided by several hypotheses about why intermediate layers can outperform final-layer representations as shown in \cref{tab:hypotheses}.
Our choices are guided by several hypotheses about why intermediate-layer representations can outperform final-layer representations:
\begin{enumerate}[label=(\roman*),noitemsep,topsep=0pt]
    \item  Intermediate layers encode rich and diverse semantic information that is beneficial for probing tasks.
    \item Certain intermediate layers capture task-relevant features that facilitate effective probe training.
    \item Intermediate layers compress redundant information while preserving task-relevant structure.
    \item Intermediate layers exhibit meaningful statistical structure with high effective information capacity.
\end{enumerate}

% \begin{enumerate}[noitemsep,topsep=0pt]
% \item[(i)] \label{hyp:h1} Intermediate layers encode rich and diverse semantic information beneficial for downstream tasks. % rankme
% \item[(ii)] Certain intermediate layers specifically capture features relevant to particular downstream tasks. % validation loss, gradient norm, snr
% \item[(iii)] Intermediate layers effectively compress the data by removing redundant information, resulting in more useful representations for downstream tasks. % curvature, ID
% \item[(iv)] Intermediate layers capture meaningful statistical structure with higher effective information capacity.
% \end{enumerate}

Motivated by these hypotheses, we introduce a set of layer-selection criteria that form the basis for efficient and automatic layer selection and are organized into three families described below. Table~\ref{tab:hypotheses} summarizes the correspondence between these hypotheses and the layer-selection criteria. 
% considered in this work.

\subsection{Notations}
\label{subsec:preliminary}
For a LLM with layers indexed by $\ell \in \{1,\dots,L\}$, let $\mathbf{H}^{(\ell)}_i \in \mathbb{R}^{T \times d}$ denote the token-wise representations at layer $\ell$ for the $i$-th input sample $x_i$, where $T$ is the number of tokens in the input and $d$ is the representation dimensionality. We denote by $\mathbf{z}^{(\ell)}_{t,i} = \mathbf{H}^{(\ell)}_{t,i} \in \mathbb{R}^d$ the representation of the $i$-th sample extracted at layer $\ell$ and token position $t$. Collecting representations across the dataset, we form the matrix $\mathbf{Z}^{(\ell)}_t \in \mathbb{R}^{N \times d}$, where each row corresponds to one sample's representation at layer $\ell$ and position $t$, with $N$ denoting the number of samples. For notational simplicity, we omit the token-position index $t$ and specify the chosen token position in context when needed.
For representation extraction, the input to the LLM consists of the concatenation of the prompt and its corresponding generated answer.
% Unless otherwise specified, we use representations extracted at the final-token position or the last token of the first generated sentence (\cref{sec:which_position}) for language tasks, and the \texttt{[CLS]} token for vision tasks. For notational simplicity, we omit the subscript $t$ when it is clear from context.

\subsection{Information-Theoretic Criteria}
% Information-theoretic metrics measure how much information a representation encodes and how evenly that information is distributed across dimensions. Representations with variance concentrated in few directions are typically less transferable than those with a more uniform spectral spread, which tend to yield richer, more robust features for downstream tasks.
Inspired by Cover’s theorem~\citep{cover1965geometrical}, which suggests that higher-rank representations are more likely to be linearly separable, we adopt RankMe~\citep{effective_rank}, which measures the rank of embeddings and has been shown to correlate strongly with downstream linear-probing performance~\citep{rankme}. 
% By operating solely on the spectrum of the representation matrix, RankMe provides a task-agnostic criterion for predicting downstream performance for pretrained self-supervised models.

Formally, given an embedding matrix $\mathbf{Z}^{(\ell)}$, RankMe considers the singular values
$\boldsymbol{\sigma}(\mathbf{Z}^{(\ell)})$ of $\mathbf{Z}^{(\ell)}$.
The normalized spectral distribution is defined as
\begin{equation*}
p_k =
\frac{\sigma_k(\mathbf{Z}^{(\ell)})}
     {\lVert \boldsymbol{\sigma}(\mathbf{Z}^{(\ell)}) \rVert_1 + \varepsilon},
\label{eq:rankme_pk}
\end{equation*}
where $\varepsilon$ is a small constant for numerical stability.
The RankMe score is then defined as
\begin{equation*}
\mathrm{RankMe}(\mathbf{Z}^{(\ell)}) =
\exp\!\left(-\sum_{k=1}^{\min(N,d)} p_k \log p_k \right).
\label{eq:rankme_def}
\end{equation*}
We select the intermediate layers with the highest RankMe value under hypothesis (i).

\subsection{Gradient-Based Criteria}
Gradient-based criteria are directly aligned with hypothesis (ii) as they measure how well the representations from a given layer facilitate learning for the downstream task.

\textbf{Validation loss} of the trained probe over a single training run is stored as an intuitive and lightweight criterion to predict the performance of the full training. 

\textbf{Relative gradient norm (RGN)} measures the magnitude of the optimization signal relative
to the scale of the model parameters.
Let $\boldsymbol{\theta}$ denote the flattened parameters of the downstream model and
$\mathbf{g}=\nabla_{\boldsymbol{\theta}}\mathcal{L}$ denote the corresponding gradient
of the validation loss.
RGN is computed as $\mathrm{RGN}~=~{\lVert \mathbf{g} \rVert_2}/{\lVert \boldsymbol{\theta} \rVert_2}$.
% \begin{equation*}
% \mathrm{RGN}= 
% % \frac{\lVert \mathbf{g} \rVert_2}{\lVert \boldsymbol{\theta} \rVert_2}.
% {\lVert \mathbf{g} \rVert_2}/{\lVert \boldsymbol{\theta} \rVert_2}.
% \end{equation*}
It has been explored for selective fine-tuning in prior work~\citep{surgical_finetuning}, where it was found to yield generally good results. We choose intermediate layers whose representations lead to larger gradient norms under the hypothesis that they carry information that leads to a more efficient learning process.

\paragraph{Signal-to-noise ratio (SNR)} characterizes the consistency of gradients across training examples.
Let $\mathbf{g}_{ij}\in\mathbb{R}$ denote the $j$-th component of the gradient computed from the $i$-th datapoint on the validation set.
We estimate SNR as
\begin{equation*}
\mathrm{SNR}
=
\mathbb{E}_{i}\!\left[
\frac{\mathrm{Avg}_{j}\!\left(\mathbf{g}_{ij}\right)^2}
{\mathrm{Var}_{j}\!\left(\mathbf{g}_{ij}\right)}
\right]
\;\propto\;
\mathbb{E}_{i}\!\left[
\frac{\left(\sum_{j} \mathbf{g}_{ij}\right)^2}
{\sum_{j} \mathbf{g}_{ij}^2}
\right].
\end{equation*}
Previous work~\citep{surgical_finetuning} has explored the use of SNR in selective fine-tuning, though the results were not as promising as RGN. Following their strategy, we select intermediate layers with higher SNR values.

\subsection{Geometric Criteria}
Geometric criteria describe how representations are organized in latent space and provide low-cost proxies for characterizing compression and structure.
% We focus on two complementary signals: curvature \cite{curvature}, which quantifies local nonlinearity along token trajectories, and intrinsic dimension, which estimates the effective degrees of freedom of the representation manifold.

\textbf{Curvature} measures the geometric complexity of token-level representation trajectories across layers and was first proposed by \citet{curvature}.
In that work, the authors observed that token embeddings tend to become progressively ``flattened'' during training, with intermediate layers exhibiting stronger flattening effects than the final layer.
% This phenomenon suggests that intermediate representations undergo substantial compression while preserving task-relevant structure.
Subsequent studies~\cite{layer_by_layer} further demonstrated that curvature is strongly correlated with downstream performance, making it a useful geometric signal aligned with hypothesis~(iii).

To compute curvature, the hidden states at layer $\ell$ are treated as a discrete trajectory across token positions, where the difference between two adjacent states along this trajectory is given by $\mathbf{v}_{t,i} = \mathbf{H}^{(\ell)}_{t,i} - \mathbf{H}^{(\ell)}_{t-1,i}, \quad t=2,\dots,T$.
% \begin{equation}
% \mathbf{v}_t = \mathbf{H}^{(\ell)}_t - \mathbf{H}^{(\ell)}_{t-1},
% \quad t=2,\dots,T,
% \end{equation}
The turning angle between consecutive velocity vectors is
\begin{equation*}
\kappa_{t,i} =
\arccos\!\left(
\frac{\langle \mathbf{v}_{t-1,i}, \mathbf{v}_{t,i} \rangle}
{\lVert \mathbf{v}_{t-1,i} \rVert_2 \, \lVert \mathbf{v}_{t,i} \rVert_2 }
\right),
\quad t=3,\dots,T.
\end{equation*}
The per-sample curvature is defined as the mean turning angle along the trajectory, that is,
\begin{equation*}
\mathrm{Curv}^{(\ell)}(x_i) = \frac{1}{T-2}\sum_{t=3}^{T} \kappa_{t,i}.
\end{equation*}
During layer selection with curvature, we choose the layer with the smallest curvature value.

\textbf{Intrinsic dimension (ID)} measures the minimum number of features required to accurately represent the embeddings' underlying structure without significant loss of information~\citep{bennett1969intrinsic}, which is closely aligned with hypothesis (iv). A variety of ID estimators have been proposed, including maximum-likelihood–based methods~\citep{mle_ID}, GeoMLE~\citep{geomle}, and TwoNN~\citep{twonn}. In this work, we adopt the TwoNN estimator due to its simplicity and scalability: it relies on the distances to the two nearest neighbors of each point and remains computationally efficient even for large datasets and high-dimensional embeddings.
% Embeddings exhibiting low intrinsic dimensions are often associated with effective data compression and tend to carry meaningful semantic information for downstream tasks ~\cite{valeriani2023geometry}, and thus this criteria captures hypothesis (ii). 

Considering the dataset-level embedding matrix
$\mathbf{Z}^{(\ell)} \in \mathbb{R}^{N\times d}$, let $r_{i,1}$ and $r_{i,2}$ denote the Euclidean distances from
$\mathbf{z}^{(\ell)}_{i}$ to its first and second nearest neighbors among the rows of
$\mathbf{Z}^{(\ell)}$.
Following the TwoNN estimator, the distance ratio is defined as $\mu_i = r_{i,2} / r_{i,1}$, 
% \begin{equation}
% \mu_i = \frac{r_{i,2}}{r_{i,1}},
% \end{equation}
% which follows a Pareto distribution with parameter $d_{\mathrm{ID}}+1$ on $[1,+\infty)$,
% \begin{equation}
% f(\mu_i ) = d_{\mathrm{ID}}\, \mu_i^{-(d_{\mathrm{ID}}+1)}.
% \end{equation}
which follows a Pareto distribution with parameter $d_{\mathrm{ID}}+1$ on $[1,+\infty)$, with density
$f(\mu_i) = d_{\mathrm{ID}}\, \mu_i^{-(d_{\mathrm{ID}}+1)}$.
Following \citet{twonn}, the ID estimation can be reduced to a linear regression task. In practice, we use the TwoNN implementation from scikit-dimension~\citep{skdim} and compute the k-nearest neighbor search by Faiss-GPU~\citep{johnson2019billion}.

We select the layer with the highest ID motivated by prior evidence that ID positively correlates with downstream performance~\citep{earthID} and layers near the maximum ID are among the earliest to transfer effectively to downstream tasks~\citep{cheng2025emergence}.

% For layer selection, we adopt a peak-based rule with a fixed look-ahead window: we identify all local maxima of the intrinsic-dimension curve across layers and examine them from shallow to deep. A candidate peak is discarded if the curve continues to increase within the subsequent window and eventually exceeds the peak value, indicating that representational capacity is still growing beyond this layer. We select the earliest remaining peak. 

% For layer selection, we adopt a peak-based rule with a fixed forward horizon $w\in\mathbb{N}$. Let $d_{\mathrm{ID}}^{(\ell)}$ denote the TwoNN intrinsic-dimension estimate at layer $\ell$ and define the forward neighborhood $\mathcal{N}^+(\ell, w)=\{\ell+1,\dots,\min(\ell+w, L)\}$. We identify all local maxima of $\{d_{\mathrm{ID}}^{(\ell)}\}_{\ell=1}^L$ and examine them from shallow to deep. A candidate peak at layer $\ell$ is discarded if $\max_{\ell'\in\mathcal{N}^+(\ell, w)} d_{\mathrm{ID}}^{(\ell')} > d_{\mathrm{ID}}^{(\ell)}$, indicating that representational capacity continues to grow beyond $\ell$ within the horizon. We select the earliest remaining peak.

\subsection{A New Criterion: FEPoID}

\begin{figure*}[t]
\centering
\includegraphics[width=0.75\linewidth]{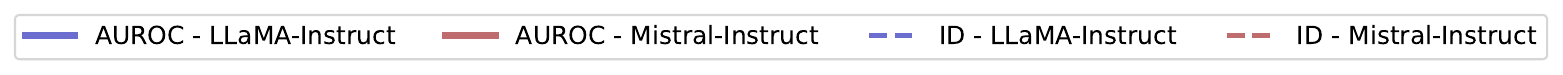}
\captionsetup[subfigure]{font=footnotesize, labelfont=footnotesize}
% 控制 tabular 默认列间距（越小越紧）
\setlength{\tabcolsep}{1pt}
\resizebox{\linewidth}{!}{%
  \begin{tabular}{@{}c@{}c@{}c@{}c@{}c@{}c@{}c@{}}
    \raisebox{0pt}[0pt][0pt]{\includegraphics[height=3.7cm]{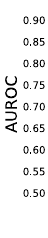}} &
    \subcaptionbox{CoQA}{\includegraphics[height=3.7cm,keepaspectratio]{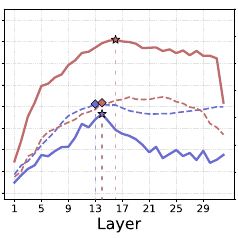}} &
    \subcaptionbox{SQuAD}{\includegraphics[height=3.7cm,keepaspectratio]{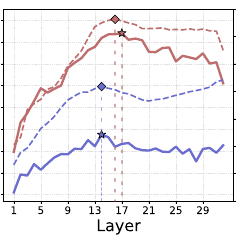}} &
    \subcaptionbox{HotpotQA}{\includegraphics[height=3.7cm,keepaspectratio]{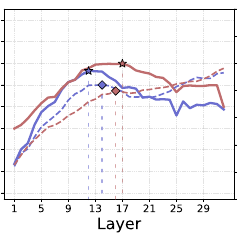}} &
    \subcaptionbox{TriviaQA}{\includegraphics[height=3.7cm,keepaspectratio]{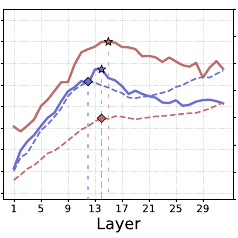}} &
    \subcaptionbox{PsiLoQA}{\includegraphics[height=3.7cm,keepaspectratio]{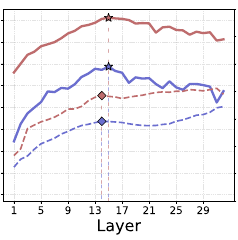}} &
    \raisebox{0pt}[0pt][0pt]{\includegraphics[height=3.7cm]{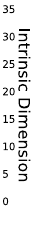}}
  \end{tabular}%
}

\caption{Layer-wise AUROC and intrinsic dimension across QA datasets. Diamond markers indicate the layers selected by FEPoID, and star markers denote the oracle best-performing layers in terms of AUROC. Across datasets and models, FEPoID consistently selects layers that are close to the oracle optima, highlighting its robustness and reliability for practical layer selection.}
\label{fig:layer_selection_no_truncation}
\end{figure*}

Through a closer examination of the evolution of the ID curves as shown in Figure~\ref{fig:layer_selection_no_truncation}, we observe a consistent multimodal pattern across models and benchmarks: 
one peak emerges in the intermediate layers, while another can appear closer to the final layers and typically attains a higher magnitude.
Although high IDs generally correlate with information-rich representations, they can arise for different underlying reasons. 
Prior work~\citep{layer_by_layer,cheng2025emergence} suggests that the intermediate layers play distinct roles in information processing, balancing the trade-off between information preservation and abstraction.

Motivated by this perspective, we hypothesize that the layer corresponding to the first peak predominantly captures the abstract semantic information, which is particularly relevant to hallucination detection. In contrast, the second peak, despite its higher magnitude, is due to the reintroduction of lexical or surface-level information as it is getting closer to predicting the next token.
%
% Based on our empirical observation that the ``early'' ID peaks often align closely with layers yielding strong probe performance, 
Based on this insight,
we propose selecting layers with the \textbf{\acf{our_method}}. 
Specifically, we use a forward horizon parameter $w$ to validate candidate peaks. Rather than naively selecting the first local maximum of the ID curve---which can be unstable when representational capacity continues to grow in deeper layers---we filter out spurious early peaks that are followed by higher ID values within a limited look-ahead window. Formally, let $d_{\mathrm{ID}}^{(\ell)}$ denote the TwoNN estimate at layer $\ell$, and define the forward horizon $\mathcal{N}^+(\ell,w)=\{\ell+1,\dots,\min(\ell+w,L)\}$. We identify all local maxima of $\{d_{\mathrm{ID}}^{(\ell)}\}_{\ell=1}^L$ and scan them from shallow to deep. A candidate peak at layer $\ell$ is discarded if 
% $\max_{\ell'\in\mathcal{N}^+(\ell,w)} d_{\mathrm{ID}}^{(\ell')} > d_{\mathrm{ID}}^{(\ell)}$
$d_{\mathrm{ID}}^{(\ell)}<d_{\mathrm{ID}}^{(\min(\ell+w,L))}$
and $d_{\mathrm{ID}}^{(\ell+1)} < d_{\mathrm{ID}}^{(\ell+2)} < \cdots < d_{\mathrm{ID}}^{(\min(\ell+w,L))}$, indicating that representational capacity continues to increase beyond $\ell$ within the horizon. We select the earliest remaining peak, defaulting to the shallowest if none survive.

% \section{Evaluation of Layer-Selection Criteria}
\section{Experiments}
% Table generated by Excel2LaTeX from sheet 'no truncation'
\begin{table*}[t]
\centering
\caption{Hallucination detection performance (AUROC) across QA datasets with $w=7$. For representation-based methods, we extract representations at the last generated token. Top-3 results are highlighted, with darker color indicating better performance.
}
\label{tab:no_truncation}
\resizebox{\textwidth}{!}{%
\begin{tabular}{cccccccccccccc}
\toprule
\multicolumn{2}{c}{}                                                                        & \multicolumn{6}{c}{LlaMA-3.1-8B-Instruct}                                                                 & \multicolumn{6}{c}{Mistral-7B-Instruct-v0.3}                                                              \\
\multicolumn{2}{c}{}                                                                        & CoQA            & SQuAD           & HotpotQA        & TriviaQA        & PsiloQA         & Avg             & CoQA            & SQuAD           & HotpotQA        & TriviaQA        & PsiloQA         & Avg             \\ \midrule
\multicolumn{2}{c}{Pred. Entropy}                                                           & 0.5833          & 0.5703          & 0.7103          & 0.6859          & 0.3604          & 0.5820          & 0.7200          & 0.7316          & 0.6303          & 0.6763          & 0.6110          & 0.6738          \\
\multicolumn{2}{c}{LN-Pred. Entropy}                                                        & 0.5781          & 0.5671          & 0.7087          & 0.6774          & 0.3555          & 0.5774          & 0.6528          & 0.6390          & 0.6589          & 0.6828          & 0.4418          & 0.6151          \\
\multicolumn{2}{c}{Semantic Entropy}                                                        & 0.5003          & 0.5518          & 0.4454          & 0.5505          & 0.6076          & 0.5311          & 0.5769          & 0.6409          & 0.6418          & 0.7353          & 0.6853          & 0.6560          \\
\multicolumn{2}{c}{Lexical Similarity}                                                      & 0.6780          & 0.5988          & 0.7294          & 0.6838          & 0.4082          & 0.6196          & 0.7071          & 0.7169          & 0.6946          & 0.7547          & 0.5781          & 0.6903          \\
\multicolumn{2}{c}{LID}                                                                     & 0.5059          & 0.5281          & 0.5171          & 0.4989          & 0.5994          & 0.5299          & 0.5274          & 0.5688          & 0.5518          & 0.4970          & 0.6447          & 0.5579          \\
\multicolumn{2}{c}{EigenScore}                                                              & 0.5247          & 0.5300          & 0.5987          & 0.5882          & 0.5080          & 0.5499          & 0.7092          & 0.7508          & 0.6587          & 0.7273          & 0.7246          & 0.7141          \\ \midrule
\multirow{7}{*}{\begin{tabular}[c]{@{}c@{}}Hidden-State\\ Probing \end{tabular}} & RankME    & \cellcolor{ranktwo}{ 0.6598}    & 0.6071          & 0.7040          & 0.7114          & 0.7478          & 0.6860          & 0.7743          & 0.7341          & 0.6909          & 0.6699          & 0.8277          & 0.7394          \\
& Curvature & \cellcolor{rankthree}{0.6323}          & \cellcolor{ranktwo}{ 0.6183}    & \cellcolor{rankthree}{0.7413}          & 0.7366          & \cellcolor{rankthree}{0.7565}          & \cellcolor{rankthree}{0.6970}         & \cellcolor{rankone}\textbf{0.8492} & \cellcolor{ranktwo}{0.8553}          & \cellcolor{rankthree}{0.7940}          & \cellcolor{ranktwo}{ 0.8368}    & \cellcolor{ranktwo}{ 0.9005}    & \cellcolor{ranktwo}{ 0.8472}    \\
& Val Loss  & \cellcolor{rankone}\textbf{0.6705} & \cellcolor{rankthree}{0.6164}          & \cellcolor{ranktwo}{ 0.7682}    & \cellcolor{rankone}\textbf{0.7861} & \cellcolor{ranktwo}{ 0.7836}    & \cellcolor{ranktwo}{ 0.7250}    & \cellcolor{rankthree}{0.8283}          & \cellcolor{rankone}\textbf{0.8679} & \cellcolor{ranktwo}{ 0.7968}    & \cellcolor{rankone}\textbf{0.8496} & \cellcolor{rankthree}{0.8861}          & \cellcolor{rankthree}{0.8457}          \\
& RGN       & 0.5993          & 0.6130          & 0.7040          & \cellcolor{ranktwo}{ 0.7859}    & 0.7373          & 0.6879          & 0.7090          & \cellcolor{rankthree}{0.7553}          & 0.7493          & \cellcolor{rankthree}{0.7868}          & 0.8562          & 0.7713          \\
& SNR       & 0.5240          & 0.5699          & 0.7009          & 0.5567          & 0.6624          & 0.6028          & 0.7475          & 0.6870          & 0.7620          & 0.7008          & 0.8207          & 0.7436          \\
& ID    & \cellcolor{rankone}\textbf{0.6705} & 0.6130          & 0.6932          & 0.7073          & 0.7373          & 0.6843          & \cellcolor{rankthree}{0.8283}          & \cellcolor{rankone}\textbf{0.8679} & 0.6993          & \cellcolor{rankthree}{0.7868}          & 0.8533          & 0.8071          \\
& FEPoID& \cellcolor{rankone}\textbf{0.6705} & \cellcolor{rankone}\textbf{0.6377} & \cellcolor{rankone}\textbf{0.7807} & \cellcolor{rankthree}{0.7516}          & \cellcolor{rankone}\textbf{0.7862} & \cellcolor{rankone}\textbf{0.7253} & \cellcolor{ranktwo}{ 0.8466}    & \cellcolor{rankone}\textbf{0.8679} & \cellcolor{rankone}\textbf{0.7982} & \cellcolor{rankone}\textbf{0.8496} & \cellcolor{rankone}\textbf{0.9031} & \cellcolor{rankone}\textbf{0.8531} \\ \bottomrule
\end{tabular}}%

\end{table*}

In this section, we investigate whether \ac{our_method} can consistently select near-optimal layers for hallucination detection across datasets and model architectures, and how it compares to widely used baselines.
In addition, we provide empirical evidence in Appendix \ref{linear_probe} to support the semantic-information hypothesis that motivates FEPoID.

\subsection{Experimental Setup}
\textbf{Dataset}
We evaluate hallucination detection across two task types: question answering (QA) and summarization.

For QA, we conduct experiments on five widely used datasets: CoQA~\citep{coqa}, SQuAD~\citep{squad}, HotpotQA~\citep{hotpotqa}, TriviaQA~\citep{triviaqa}, and PsiLoQA~\citep{psiloqa}.
CoQA, SQuAD, and PsiLoQA are evaluated in a \emph{context-aware} setting, where the input prompt includes the supporting passage with the question. HotpotQA and TriviaQA are evaluated in a \emph{question-only} setting.
Following \citet{illusion_of_progress,semantic_entropy_nature}, we sample 10 candidate answers per question at a temperature of 1.0 to quantify generation uncertainty; these samples are used by baselines that rely on multiple sampled outputs.
In addition, we generate a single \emph{best answer} using a temperature of 0.1 for each question, which serves as a deterministic estimate for downstream performance evaluation. 
To assess the layer selection criteria beyond QA tasks, we further evaluate on two summarization benchmarks: HaluEval~\citep{li2023halueval} and CNN/Daily Mail (CNN/DM)~\citep{cnn_daily}, in which the goal is to classify whether the LLM summarizes correctly. 

For each QA dataset, answers are generated autoregressively with a maximum generation length of 30 tokens. For summarization datasets, the maximum generation length is set to 130. Details on prompt templates and dataset construction are provided in Appendix~\ref{app:HD_SM}.
\paragraph{Model}
We conduct experiments across a diverse set of models varying in size (1B--8B), tuning strategy (base vs.\ instruction-tuned), and architecture. For instruction-tuned models, we experiment with \acl{llama}~\citep{grattafiori2024llama} (abbreviated as \textbf{\ac{llama}}) and \acl{mistral}~\citep{jiang2023mistral7b} (abbreviated as \textbf{\ac{mistral}}). To assess robustness across tuning strategies, we additionally evaluate on LLaMA-3.1-8B (base). To further assess scalability, we also include LLaMA-3.2-1B and LLaMA-3.2-3B.

\textbf{Evaluation Setup}
Hallucination detection is a binary classification task. In QA tasks, for each generated answer, we first compare it against the reference answer using exact string matching. If there is an exact match, the answer is labeled as correct. If not, we use an LLM-as-Judge to assign the label, following the prompt and procedure of \citet{llmknows}.
For summarization tasks, the ground-truth labels indicating whether the LLM summarizes correctly are constructed via TrueTeacher~\citep{gekhman2023trueteacher}. We report AUROC on the test split for each dataset.

\textbf{Hallucination Detection Baseline}
While our main focus is the layer-selection problem, we also compare hallucination detection performance against widely used baselines to contextualize our results.

We first consider several uncertainty-based baselines. Predictive Entropy (Pred. Entropy) and Length-Normalized Predictive Entropy (LN-Pred. Entropy)~\citep{predictive_entropy_and_ln_entropy} quantify uncertainty by measuring the variability of the model's likelihood across multiple sampled generations. We additionally include Semantic Entropy~\citep{semantic_entropy_nature}, which estimates uncertainty at the semantic level by clustering sampled answers into equivalence classes and measuring their consistency. Complementary to these uncertainty-based approaches, we include Lexical Similarity~\citep{lin2024generating} as a surface-form baseline, which measures token-level overlap between the generated answer and the reference using ROUGE-L.

Finally, we evaluate representation-based baselines, including EigenScore~\citep{eigenscore} and Local Intrinsic Dimension (LID)~\citep{LID}. EigenScore assesses representation quality via the spectral properties of hidden-state covariance and is applied to middle-layer representations. LID estimates local intrinsic dimensionality under the hypothesis that truthful outputs exhibit more structured representations; following \citet{LID}, we probe the layer immediately after the maximum-LID layer.
% SAPLMA trains probing classifiers (MLPs) on layer-wise hidden representations of the language model, using the hidden state of the last token, to predict answer correctness. Throughout all probing experiments, the underlying language model is kept frozen, ensuring that performance differences arise from representational properties rather than finetuning effects. the structure of MLP, training parameters and procedure please \textit{refer to appendix}.

\textbf{Hidden-State Probing Setup}
% We use a standard hidden-state probing setup to assess how layer representations support hallucination detection, keeping the language model frozen so performance differences reflect representation quality rather than finetuning.
%
% For RankMe, curvature, and ID–based methods (max ID and \ac{our_method})—we first compute the corresponding metric using the hidden representations at a fixed token position at each layer. Each criterion then selects a single preferred layer based on its own rule. An MLP probe is subsequently trained and evaluated using representations extracted from the selected layer.
%
% In contrast, for training-dynamics criteria (validation loss, RGN, SNR), we train an MLP probe independently for \emph{each} layer. Validation loss, RGN, and SNR are then computed on the validation set and used as layer-selection signals.
We train a lightweight MLP at each layer using a fixed token position (specified in later ablations) and select the checkpoint with the lowest validation loss. Curvature, RankMe, and ID are computed on both the training and validation sets, while validation loss, RGN, and SNR are computed on the validation set only. 
% Each criterion selects a single layer according to its own rule.

\subsection{Empirical Results of \ac{our_method}}
\begin{figure}[t]
    \centering
    \includegraphics[width=0.9\linewidth]{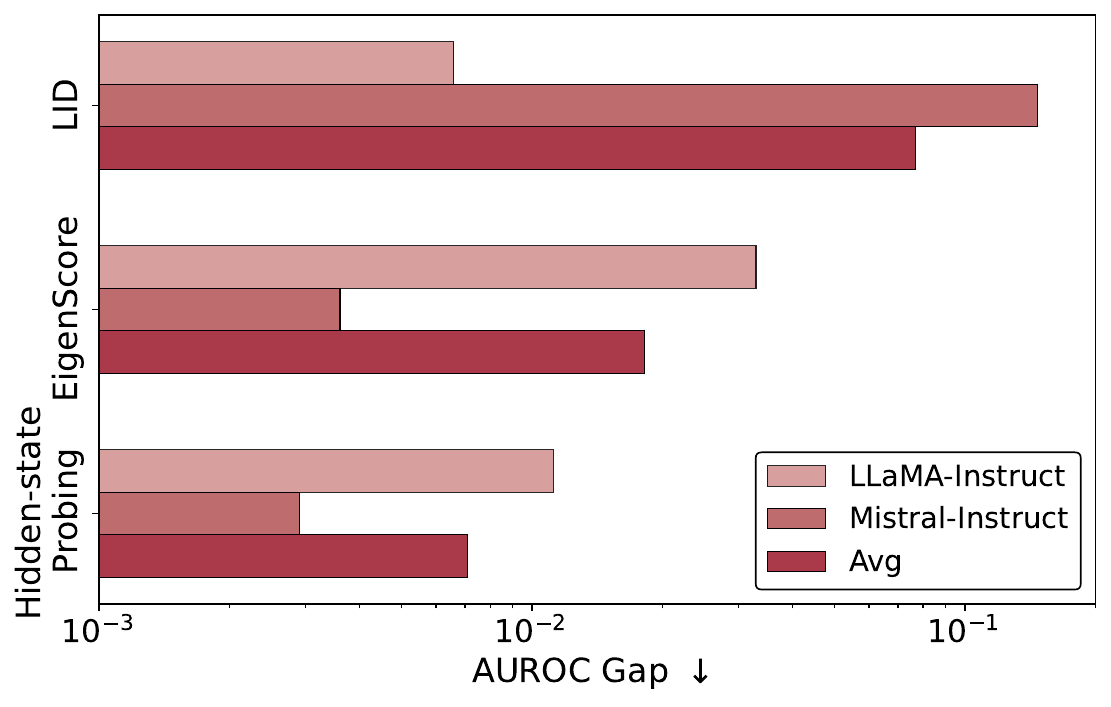}
    \caption{AUROC gap between the layer selected by each method and the oracle best-performing layer. \textbf{LLaMA-Instruct} and \textbf{Mistral-Instruct} denote model-specific averages over datasets, while \textbf{Avg} further averages across all models and datasets. For hidden-state probing, layers are selected by \ac{our_method}.}
    \label{fig:auroc_gap}
\end{figure}

\paragraph{QA Task}
\cref{tab:no_truncation} reports AUROC across five QA datasets and two instruction-tuned LLMs. For both probe training and layer-selection criteria, we use the hidden state at the last generated token. Overall, combining hidden-state probing with \ac{our_method} yields the strongest performance among all baselines and selection criteria. This result indicates that \ac{our_method} more accurately and stably identifies layers whose representations support highly discriminative probes compared to alternative layer-selection strategies.
\label{sec:which_position}
\begin{figure*}[t]
    \centering
    \includegraphics[width=0.98\textwidth]{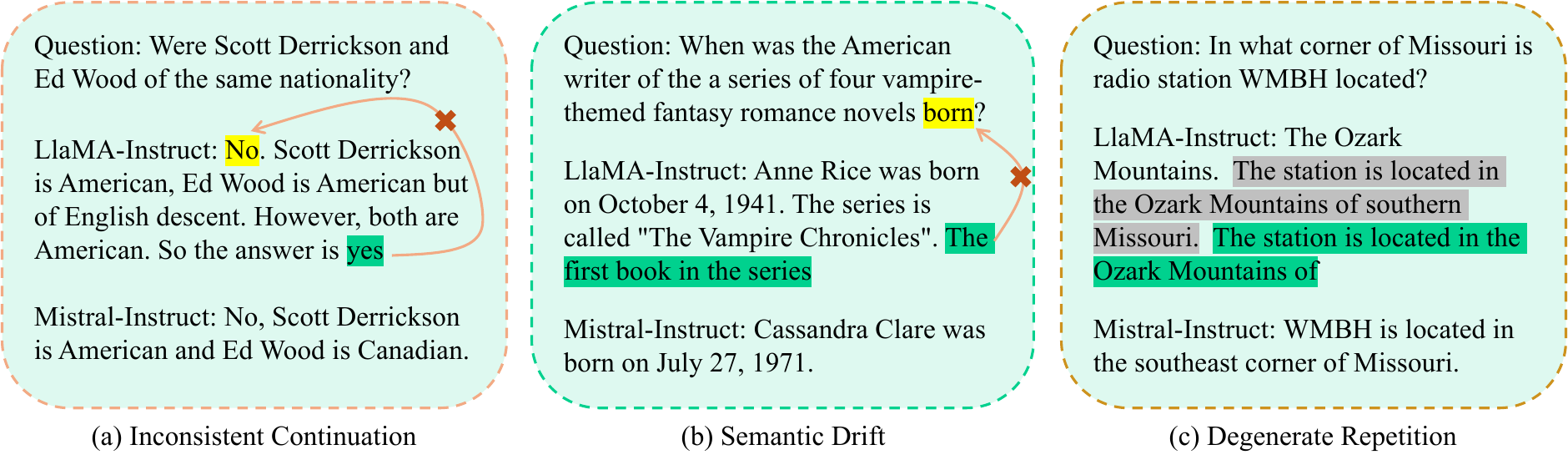}
    \caption{A comparison of generation behaviors in LLaMA-Instruct and Mistral-Instruct without FST. Specifically, (a) shows an internally inconsistent continuation where \ac{llama} contradicts its initial answer, (b) demonstrates semantic drift in which the generation deviates from the question focus, and (c) highlights degenerate repetition with redundant restatement of the same information. In contrast, Mistral-Instruct consistently produces concise and well-terminated responses.}
    \label{fig:different_behavior}
\end{figure*}
\begin{table}[t]
\centering
\caption{Results (AUROC) on \textbf{summarization tasks} with $w = 7$, without FST. FEPoID outperforms all other criteria, demonstrating its effectiveness beyond QA tasks. Notably, Val Loss \textit{fails} to rank among the top-2 criteria on all settings.}
\label{tab:summarization}

\resizebox{\columnwidth}{!}{%
\begin{tabular}{ccccccc}
\toprule
          & \multicolumn{3}{c}{LlaMA-3.1-8B-Instruct}                                                                     & \multicolumn{3}{c}{Mistral-7B-Instruct-v0.3}                                                                 \\
          & HaluEval                        & CNN/DM                   & Avg                                & HaluEval                       & CNN/DM                 & Avg                                \\ \midrule
RankMe    & \cellcolor{ranktwo}0.6075           & 0.5774                             & \cellcolor{rankthree}0.5924        & 0.7149                             & 0.6869                             & 0.7009                             \\
Curvature & 0.5494                              & \cellcolor{ranktwo}0.5922          & 0.5708                             & \cellcolor{rankthree}0.7498        & \cellcolor{ranktwo}0.7319          & \cellcolor{ranktwo}0.7409          \\
Val Loss  & \cellcolor{rankthree}0.5961         & 0.5859                             & 0.5910                             & 0.7294                             & 0.6938                             & 0.7116                             \\
RGN       & 0.5713                              & 0.5821                             & 0.5767                             & \cellcolor{ranktwo}0.7563          & 0.7031                             & 0.7297                             \\
SNR       & 0.5528                              & 0.5474                             & 0.5501                             & 0.7385                             & 0.6811                             & 0.7098                             \\
ID        & \cellcolor{ranktwo}0.6075           & \cellcolor{rankthree}0.5918        & \cellcolor{ranktwo}0.5997          & \cellcolor{rankthree}0.7498        & \cellcolor{rankthree}0.7185        & \cellcolor{rankthree}0.7342        \\
FEPoID    & \cellcolor{rankone}\textbf{0.6165 } & \cellcolor{rankone}\textbf{0.5995} & \cellcolor{rankone}\textbf{0.6080} & \textbf{\cellcolor{rankone}0.7808} & \textbf{\cellcolor{rankone}0.7614} & \textbf{\cellcolor{rankone}0.7711} \\ \bottomrule
\end{tabular}
}
\end{table}
As shown in \cref{tab:no_truncation}, different selection rules applied to the hidden-state probing framework can lead to markedly different outcomes across models and datasets. In particular, suboptimal layer selection may significantly degrade performance: for example, when using SNR to select layers for \ac{llama}, the average resulting AUROC is lower than that of simple baselines such as lexical similarity. In addition, \cref{fig:auroc_gap} further quantifies the discrepancy between the layer selected by each method and the best-performing layer for LID, EigenScore, and hidden-state probing, all of which are representation-based approaches. We observe that LID and EigenScore incur relatively large AUROC gaps, indicating that their layer-selection strategies fail to reliably identify informative layers. In contrast, layers selected by \ac{our_method} in the hidden-state probing framework yield substantially smaller gaps, remaining close to the best-performing layers across models and datasets. These results further demonstrate that effective selection criteria are crucial for fully realizing the benefits of representation-based probing.

% Among all layer selection strategies, \ac{our_method} achieves the most robust and consistently strong performance. On both LLaMA-3.1-8B-Instruct and Mistral-7B-Instruct-v0.3, FEPoID attains the highest average AUROC across datasets, and is either the best or tied for the best method on nearly all individual benchmarks. Compared to alternatives such as validation loss, curvature, or signal-to-noise–based criteria, FEPoID generalizes more reliably across models and datasets without relying on training dynamics or task-specific supervision.

Notably, selecting the layer with the maximum ID does not consistently yield optimal performance, as illustrated in \cref{fig:layer_selection_no_truncation}. In datasets such as HotpotQA, TriviaQA, and PsiLoQA, the maximal ID often appears in very late layers, where representations become overly complex or redundant for hallucination detection, whereas the first effective peak stays close to the best-performing layer. 

\textbf{Summarization Tasks} 
We extend the hallucination detection experiments further to summarization tasks. As shown in \cref{tab:summarization}, FEPoID achieves the best performance across all datasets on both models, demonstrating its robustness beyond QA tasks. Notably, Val Loss consistently fails to rank among the top-3 criteria in terms of average AUROC on both models, suggesting that validation-based criteria are less reliable in summarization tasks.

\textbf{Sensitivity to Hyperparameters} \cref{fig:w} presents an ablation study on the forward horizon size $w$ used in \ac{our_method}. The performance of \ac{our_method} remains highly stable across a wide range of $w$ for all datasets and models, indicating that the method is highly robust to the choice of $w$.

\textbf{Time Efficiency of FEPoID} We report the computation time (in seconds) of each criterion on LLaMA-3.1-8B-Instruct in \cref{tab:time}. Note that the time for Val Loss includes both MLP training and validation loss computation. FEPoID (equivalently, ID) achieves the lowest computation time across all benchmarks, demonstrating the computational \emph{efficiency} of FEPoID.
\begin{table}[t]
\centering
\caption{Computation time (in seconds) of each criterion on 
LlaMA-3.1-8B-Instruct, measured as the total time across all 32 layers. 
FEPoID and ID require significantly less computation time than all other criteria.}
\label{tab:time}
\resizebox{\columnwidth}{!}{%
\begin{tabular}{ccccccc}
\toprule
           & CoQA  & SQuAD & HotpotQA & TriviaQA & PsiLoQA & Avg   \\ \midrule
RankMe     & 26.32 & 27.65 & 28.98    & 28.17    & 25.41   & 27.30 \\
Curvature  & 43.88 & 45.23 & 45.55    & 46.16    & 45.37   & 45.24 \\
Val Loss   & 27.59 & 30.81 & 29.80    & 29.79    & 29.78   & 29.55 \\
RGN        & 54.83 & 60.95 & 58.31    & 58.08    & 58.72   & 58.18 \\
SNR        & 54.20 & 59.91 & 57.31    & 58.84    & 59.16   & 57.88 \\

FEPoID; ID & \textbf{9.40}  & \textbf{9.90}  & \textbf{9.95}     & \textbf{10.59 }   & \textbf{10.88}   & \textbf{10.14} \\ \bottomrule
\end{tabular}
}
\end{table}
\paragraph{Generalization Across Scales and Tuning Strategies}
We evaluate FEPoID on base models and other-scale ones. As shown in \cref{tab:model_8B_base,tab:model_scales}, FEPoID achieves the highest AUROC on 4 out of 5 datasets and the highest average AUROC on LLaMA-3.1-8B (base), and consistently ranks among the top-performing criteria on LLaMA-3.2-3B and LLaMA-3.2-1B, demonstrating its generalization beyond instruction-tuned settings and across varying model scales.

\section{Which Token Position Should We Probe? }

In this section, we study which token position should be probed for hidden-state probing. We show that representations at the last generated token are often degraded by end-of-sequence noise, and propose \textbf{first-sentence truncation (FST)} as a simple, supervision-free remedy. Our experiments show that FST yields  more discriminative class structure in the extracted representations, and consistently improves all hallucination detection methods considered.

\subsection{First-Sentence Truncation}
Beyond layer selection, the effectiveness of hidden-state probing also critically depends on the \emph{token position} $t$ at which the representation $\mathbf{z}^{(\ell)}_{t,i}$ is extracted. 
In practice, a common heuristic extracts representations at the last generated token, i.e., $t=T$, motivated by the autoregressive property that this token can attend to the entire preceding context. However, as illustrated in \cref{fig:different_behavior}, the representations extracted at the final token position are frequently corrupted by end-of-sequence noise, arising from inconsistent continuations, semantic drift, and degenerate repetition.

Prior work~\citep{llmknows} demonstrates that probing hidden states aligned with the ``exact answer'' tokens (i.e., the token in the generated output that directly matches the ground-truth answer) yields substantially stronger performance. However, identifying such tokens requires access to ground-truth answers and often an auxiliary LLM, which is impractical in real-world applications.

Motivated by the empirical observation that LLMs often state the answer early in the generation---typically within the first sentence---and by prior findings that the truthfulness information is concentrated at answer-aligned tokens~\citep{llmknows}, we extract the representations at the last token of the first generated sentence as a lightweight approximation to answer-aligned representations. Specifically, we set $t$ to the index of the last token of the first generated sentence for each sample $x_i$, and extract $\mathbf{z}^{(\ell)}_{t,i}$ accordingly. Compared to setting $t=T$, this choice is less susceptible to end-of-sequence noise.
In addition, unlike extracting representations at the exact-answer positions, this strategy requires neither access to ground-truth answers nor auxiliary LLMs, making it a practical and efficient alternative for real-world applications.

To identify sentence boundaries, we employ a lightweight, rule-based scanner to perform \textbf{first-sentence truncation (FST)}. 
% Implementation details are provided in Appendix~\ref{app:implementation_FST}.
Implementation details are in Appendix~\ref{app:implementation_FST}.

% \section{FST Helps All Baselines}
\subsection{Empirical Evaluation of FST}
\begin{figure}[t]
    \centering
    \includegraphics[width=0.9\linewidth]{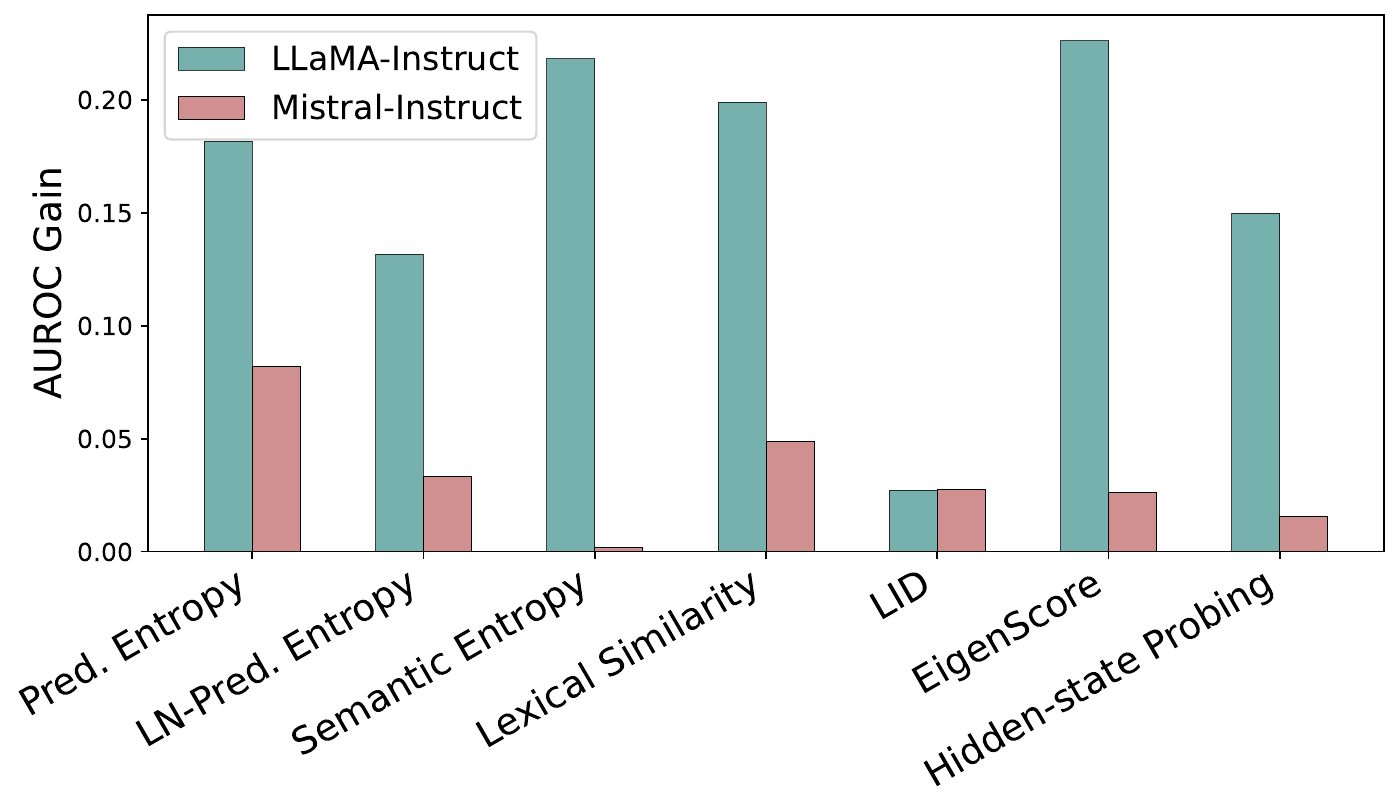}

    \caption{AUROC improvements obtained by applying FST relative to the ``last generated token'' heuristic for each method, averaged over datasets. The layers for the hidden-state probing framework are selected by \ac{our_method} with $w=7$.}
    \label{fig:fst_ablation}
\end{figure}

% \begin{figure}[t]
%     \centering
%     \includegraphics[width=0.85\linewidth]{figures/specific_example.png}
%     \caption{Different Generation Behaviors of Two Models. Mistral-Instruct terminates generation earlier and produces a concise and consistent response.}
%     \label{fig:generation_behavior}
% \end{figure}
% As discussed in \cref{sec:which_position}, 
We compare two token positions for feature extraction: (i) the last token of the generated sequence and (ii) the last token of the first generated sentence, identified using FST.
% two strategies for hallucination detection: (i) the default “Last Token” heuristic, which uses the hidden state of the last generated token, and (ii) using the hidden state of the last token in the first generated sentence. 
% To identify sentence boundaries, we employ a lightweight, rule-based scanner to operate First-Sentence Truncation (\textbf{FST}). Specifically, the scanner processes the generated text from left to right and identifies the first period that does not fall under common exception cases as the end of the first sentence. These exceptions include ellipses (\texttt{…}), decimal numbers (e.g., \texttt{3.14}), multi-dot abbreviations (e.g., \texttt{U.S.}, \texttt{i.e.}), word-level abbreviations (e.g., \texttt{Dr.}, \texttt{etc.}, \texttt{No.\ 3}), and name initials (e.g., \texttt{G. Smith}). The scanner is simply implemented with regular expression operations.

After applying FST to each candidate answer and each best-answer, we extract the hidden states at the last token of the truncated answers for LID, EigenScore, and hidden-state probing. For uncertainty-based baselines and lexical similarity, we evaluate them on the first-sentence truncated candidate answers to ensure consistency across methods.

\cref{fig:fst_ablation} shows that FST delivers consistent AUROC gains across all methods. These results show that FST benefits multiple levels of information used for hallucination detection, including representation-level signals, likelihood-based signals (uncertainty-based baselines), and surface-form signals (Lexical Similarity). By truncating generations at the end of the first sentence, FST removes interfering effects introduced by later-stage generation, thereby stabilizing diverse signals in a method-agnostic manner rather than benefiting any single assumption or criterion.
% This corroborates prior evidence that last-token representations of the whole generated sequence are sensitive to end-of-sequence noise \cite{springer2025repetition,lee2025nvembed} and further 
% demonstrating the effectiveness and practical necessity of FST. 

Notably, as shown in \cref{fig:fst_ablation}, FST yields substantially larger improvements in hallucination detection for \ac{llama} than for \ac{mistral}. This discrepancy is largely driven by differences in generation behavior: \ac{mistral} typically emits an \texttt{<eos>} token shortly after completing the first sentence, whereas \ac{llama} often continues generating tokens until reaching the configured maximum length limit. As a result, representations extracted at the final generated token in \ac{llama} are more likely to be contaminated by noisy continuations. We show three specific examples in \cref{fig:different_behavior}, where the extra continuation in \ac{llama} introduces end-of-sequence noise in several recurring forms: (a) \emph{inconsistent continuation}, where the model initially states an answer but later produces a conflicting one; (b) \emph{semantic drift}, where the continuation shifts into content unrelated to the question (e.g., adding irrelevant details after answering); and (c) \emph{degenerate repetition}, where the model repeats phrases without adding information. These noisy continuations corroborate prior evidence that last-token representations extracted from the entire generated sequence are degraded by end-of-sequence noise and therefore tend to perform poorly on downstream tasks~\citep{springer2025repetition,lee2025nvembed}.

We report the detection results with FST in \cref{tab:first_sentence_truncation} and the layer-wise AUROC and ID in \cref{fig:layer_selection_FST}. The overall trends closely mirror those observed in \cref{tab:no_truncation}, with \ac{our_method} consistently achieving the strongest performance across models.
\paragraph{Sensitivity to Hyperparameters} When FST is applied, \ac{our_method} remains robust to the choice of $w$ (\cref{fig:w}). The only exception is TriviaQA, where larher $w$ leads to visibly stronger performance. 

\textbf{Generalizability of FST across Models} We further apply FST to base and other-scale models, and report the results in \cref{tab:model_scales,tab:model_8B_base}. Compared with the evaluation without FST in \cref{tab:model_scales_without_FST,tab:model_8B_base_without_FST}, FST consistently enhances the hallucination detection performance, demonstrating its  generalizability across model scales and tuning strategies.

\begin{table}[t]
\centering
\caption{Fisher Separation and Silhouette Score with and without FST on LLaMA-3.1-8B-Instruct. FST consistently improves both metrics across datasets, indicating that FST yields more separable representations for hallucination detection ($\uparrow$ higher is better).}
\label{tab:fst_separation}
\footnotesize
\begin{tabular}{ccccc}
\toprule
\multirow{2}{*}{Dataset} & \multicolumn{2}{c}{Fisher Separation $\uparrow$} & \multicolumn{2}{c}{Silhouette Score $\uparrow$} \\
                         & w/o FST            & w/ FST           & w/o FST           & w/ FST           \\ \hline
CoQA                     & 0.0003             & 0.0044           & -0.0007           & 0.0324           \\
SQuAD                    & 0.0004             & 0.0049           & 0.0006            & 0.0973           \\
HotpotQA                 & 0.0009             & 0.0075           & -0.0001           & -0.0011          \\
TriviaQA                 & 0.0013             & 0.0185           & 0.0059            & 0.1006           \\
PsiLoQA                  & 0.0016             & 0.0340           & 0.0118            & 0.2605           \\ \bottomrule
\end{tabular}
\end{table}

\subsection{Empirical Analysis of FST} 

Notably, after applying FST, the ID of representations extracted from the selected layers does not change substantially (as shown in \cref{fig:layer_selection_no_truncation,fig:layer_selection_FST}), while hallucination detection performance improves markedly. 
% This suggests that FST does not push representations into higher-dimensional spaces, but instead reorganizes their geometric structure in a way that is more conducive to hallucination detection.

To further investigate why FST helps, we measure class separability of the representations at the selected layers via Fisher Separation~\citep{fisher1936use} (between- vs. within-class variance) and Silhouette Score~\citep{rousseeuw1987silhouettes} (per-sample cohesion/separation), capturing both global and local aspects of class separability. As shown in \cref{tab:fst_separation}, FST consistently improves both metrics across datasets. These findings suggest that, despite residing in spaces of similar intrinsic dimensionality, representations extracted with FST exhibit cleaner and more discriminative class structure, making them more conducive to hallucination detection.
\section{Conclusion and Future Directions}

This work studies the problem of automatic layer selection for hallucination detection under the hidden-state probing framework. We conduct a systematic evaluation of a diverse set of layer-selection criteria, spanning information-theoretic, geometric, and gradient perspectives. Our results show that criteria previously shown to correlate with downstream performance or to support selective fine-tuning do not reliably yield effective layer-selection results in this setting. This highlights the gap between representation-level analysis and practical layer-selection strategies.

Based on empirical regularities in ID trajectories across layers, we propose FEPoID, a lightweight and training-free criterion that selects the first effective peak of ID. Evaluated on hallucination detection benchmarks spanning both QA and summarization tasks, across model architectures, scales, and tuning strategies, FEPoID consistently selects high-performing layers and outperforms all baselines, demonstrating its robustness and broad applicability.

We further revisit token-position choices in decoder-only LLMs and show that representations extracted at the last generated token are often degraded by end-of-sequence noise. To address this, we propose to probe at the last token of the first generated sentence via a simple, rule-based truncation, which yields consistent performance gains across all hallucination detection baselines considered. We further analyze why FST works, showing that, despite residing in spaces of similar intrinsic dimensionality, representations extracted with FST exhibit cleaner and more discriminative class structure, as evidenced by consistent improvements in Fisher Separation and Silhouette Score across datasets.

Together, FEPoID and FST offer a practical, supervision-free solution for robust representation extraction, requiring neither model fine-tuning nor exhaustive layer-wise validation. Future work may extend these findings to tasks that require abstract information processing and to different data modalities, and further investigate the theoretical relationship between ID dynamics and task-relevant representations.

\clearpage
\section*{Acknowledgments}
We would like to thank Daniel Bai and Fan Yin for their helpful discussions.
The authors acknowledge the Research Computing at the University of Virginia.
This work is funded in part by NSF CAREER IIS-2145492 and DARPA AIQ HR00112590066.

\section*{Impact Statement}
This paper presents work whose goal is to increase understanding of deep learning, which may lead to advancements in the field of Machine Learning. There are many potential societal consequences of our work, none of which we feel must be specifically highlighted here.

\bibliography{main}
\bibliographystyle{icml2026}

%%%%%%%%%%%%%%%%%%%%%%%%%%%%%%%%%%%%%%%%%%%%%%%%%%%%%%%%%%%%%%%%%%%%%%%%%%%%%%%
%%%%%%%%%%%%%%%%%%%%%%%%%%%%%%%%%%%%%%%%%%%%%%%%%%%%%%%%%%%%%%%%%%%%%%%%%%%%%%%
% APPENDIX
%%%%%%%%%%%%%%%%%%%%%%%%%%%%%%%%%%%%%%%%%%%%%%%%%%%%%%%%%%%%%%%%%%%%%%%%%%%%%%%
%%%%%%%%%%%%%%%%%%%%%%%%%%%%%%%%%%%%%%%%%%%%%%%%%%%%%%%%%%%%%%%%%%%%%%%%%%%%%%%
\newpage

\appendix

% \crefname{subsubsection}{Appendix}{Appendices}
% \Crefname{subsubsection}{Appendix}{Appendices}
\onecolumn
\section{Experimental Details}
\label{app:HD_SM}

\paragraph{QA Task}We evaluate our methods on five question answering datasets: CoQA, SQuAD, HotpotQA, TriviaQA, and PsiLoQA.  
For each dataset, we construct a training set of 9{,}000 examples and a validation set of 1{,}000 examples, which are used for probe training and hyperparameter selection.  
The test sets are kept fixed and are used exclusively for evaluation. Specifically, the test set sizes are 7{,}983 for CoQA, 10{,}000 for SQuAD, 7{,}405 for HotpotQA, 10{,}000 for TriviaQA, and 8{,}103 for PsiLoQA. 
\paragraph{Summarization Task}
We evaluate on two summarization datasets: CNN/DailyMail and HaluEval. For CNN/DailyMail, we construct a training set of 9{,}000 examples and a validation set of 1{,}000 examples, with a test set of 10{,}000 examples. For HaluEval, we use 7{,}200 training examples, 800 validation examples, and 2{,}000 test examples.
\paragraph{Prompt Template}
To generate answers for hallucination detection, we use two prompt settings depending on whether a supporting passage is available. \cref{tab:prompt_templates} summarizes the templates used in the question-only setting (HotpotQA, TriviaQA) and the context-aware setting (CoQA, SQuAD, PsiLoQA, HaluEval, CNN/Daily ), where \texttt{[Q]} and \texttt{[C]} denote the question and context, respectively.
\paragraph{Implementation of First-Sentence Truncation}
\label{app:implementation_FST}
First-sentence truncation is implemented via a rule-based scanner. Specifically, the scanner processes the generated text from left to right and identifies the first period that does not fall under common exception cases as the end of the first sentence. These exceptions include ellipses (\texttt{…}), decimal numbers (e.g., \texttt{3.14}), multi-dot abbreviations (e.g., \texttt{U.S.}, \texttt{i.e.}), word-level abbreviations (e.g., \texttt{Dr.}, \texttt{etc.}, \texttt{No.\ 3}), and name initials (e.g., \texttt{G. Smith}). The scanner is simply implemented with regular expression operations.
\begin{table}[b]
\centering
\small
\caption{Prompt templates used for answer generation.
Here, [Q] denotes the question text and [C] denotes the provided context passage.}
\label{tab:prompt_templates}
\begin{tabular}{lcl}
\toprule
\textbf{}             & Dataset                                                               & Prompt Template                                                                                                                                                 \\ \midrule
Question-Only Setting & \begin{tabular}[c]{@{}c@{}}HotpotQA \\ TriviaQA\end{tabular}          & \begin{tabular}[c]{@{}l@{}}Answer the question as briefly as possible, using plain text only:\\ Question: {[}Q{]}\\ Answer:\end{tabular}                        \\ \midrule
Context-Aware Setting & \begin{tabular}[c]{@{}c@{}}CoQA \\ SQuAD\\ PsiLoQA\end{tabular}       & \begin{tabular}[c]{@{}l@{}}Answer the question as briefly as possible, based only on the context:\\ Context: {[}C{]}\\ Question: {[}Q{]}\\ Answer:\end{tabular} \\ \midrule
Summarization Tasks   & \begin{tabular}[c]{@{}c@{}}HaluEval\\ CNN/Daily Mail\end{tabular} & \begin{tabular}[c]{@{}l@{}}Summarize the following document in one or two concise sentences.\\ Document: {[}C{]}\\Summary:\end{tabular}                           \\ \bottomrule
\end{tabular}
\end{table}

\begin{table}[]
\centering
\caption{Results on the \textbf{LlaMA-3.1-8B base} model with $w=7$. The representations are extracted without FST. FEPoID maintains strong performance on a non-instruction-tuned 
model, demonstrating that our method \emph{generalizes} across both base and 
instruction-tuned models. }
\label{tab:model_8B_base_without_FST}
\begin{tabular}{ccccccc}
\toprule
          & CoQA                               & SQuAD                              & HotpotQA                           & TriviaQA                           & PsiloQA                            & Avg                                \\ \midrule
RankME    & 0.6972                             & 0.7132                             & 0.6458                             & 0.6472                             & 0.8032                             & 0.7013                             \\
Curvature & \cellcolor{rankthree}0.7187        & \cellcolor{rankthree}0.7432        & \textbf{\cellcolor{rankone}0.7370} & \cellcolor{rankthree}0.7846        & \cellcolor{ranktwo}0.8639          & \cellcolor{rankthree}0.7695        \\
Val Loss  & \textbf{\cellcolor{rankone}0.7552} & \cellcolor{ranktwo}0.7811          & \cellcolor{ranktwo}0.7326          & \textbf{\cellcolor{rankone}0.8141} & \cellcolor{rankthree}0.8434        & \cellcolor{ranktwo}0.7853          \\
RGN       & 0.6859                             & 0.7313                             & \cellcolor{rankthree}0.6955        & 0.7257                             & 0.7816                             & 0.7240                             \\
SNR       & 0.5118                             & 0.6387                             & 0.6115                             & 0.6304                             & 0.7677                             & 0.6320                             \\
ID        & \cellcolor{ranktwo}0.7468          & \textbf{\cellcolor{rankone}0.7892} & 0.6672                             & 0.6179                             & \textbf{\cellcolor{rankone}0.8655} & 0.7373                             \\
FEPoID    & \cellcolor{ranktwo}0.7468          & \textbf{\cellcolor{rankone}0.7892} & \cellcolor{ranktwo}0.7326          & \cellcolor{ranktwo}0.8136          & \textbf{\cellcolor{rankone}0.8655} & \textbf{\cellcolor{rankone}0.7895} \\ \bottomrule
\end{tabular}
\end{table}

\begin{table}[]
\centering
\caption{Results on different model scales, where representations are extracted without FST. Forward horizon is set to $w=7$ for \textbf{LlaMA-3.2-3B} and $w = 3$ for \textbf{LlaMA-3.2-1B}. Top-3 results are highlighted, with darker color indicating better performance. FEPoID achieves consistently strong performance across both model scales, demonstrating its \emph{generalizability} to models of varying scales. 
}
\label{tab:model_scales_without_FST}
\footnotesize
\resizebox{\textwidth}{!}{%
\begin{tabular}{ccccccccccccc}
\toprule
          & \multicolumn{6}{c}{LlaMA-3.2-3B}                                                                                                                                                                                            & \multicolumn{6}{c}{LlaMA-3.2-1B}                                                                                                                                                                                            \\
          & CoQA                               & SQuAD                              & HotpotQA                           & TriviaQA                           & PsiloQA                            & Avg                                & CoQA                               & SQuAD                              & HotpotQA                           & TriviaQA                           & PsiloQA                            & Avg                                \\ \midrule
RankME    & 0.5981                             & 0.6476                             & 0.6914                             & \cellcolor{rankthree}0.6639        & 0.7469                             & 0.6696                             & 0.5621                             & 0.5902                             & 0.6583                             & 0.5963                             & 0.6441                             & 0.6102                             \\
Curvature & \cellcolor{ranktwo}0.7123          & \cellcolor{rankthree}0.6616        & \cellcolor{rankthree}0.7268        & \cellcolor{ranktwo}0.6734          & \cellcolor{rankthree}0.8280        & 0.7204                             & \cellcolor{rankthree}0.6699        & \cellcolor{ranktwo}0.6442          & \cellcolor{ranktwo}0.7587          & \cellcolor{rankone}0.7028          & \cellcolor{rankthree}0.7355        & \cellcolor{rankthree}0.7022        \\
Val Loss  & \textbf{\cellcolor{rankone}0.7391} & \cellcolor{ranktwo}0.7153          & \textbf{\cellcolor{rankone}0.7439} & \textbf{\cellcolor{rankone}0.7198} & \textbf{\cellcolor{rankone}0.8498} & \cellcolor{ranktwo}0.7536          & \textbf{\cellcolor{rankone}0.6860} & \cellcolor{ranktwo}0.6442          & \textbf{\cellcolor{rankone}0.7641} & \cellcolor{ranktwo}0.6985          & \textbf{\cellcolor{rankone}0.7597} & \textbf{\cellcolor{rankone}0.7105} \\
RGN       & \cellcolor{rankthree}0.6369        & 0.6567                             & 0.6914                             & \textbf{\cellcolor{rankone}0.7198} & 0.7990                             & 0.7008                             & 0.5777                             & 0.5902                             & \cellcolor{rankthree}0.7370        & \cellcolor{rankthree}0.6050        & 0.6597                             & 0.6339                             \\
SNR       & 0.5116                             & 0.5544                             & \textbf{\cellcolor{rankone}0.7439} & 0.5748                             & 0.6878                             & 0.6145                             & 0.6226                             & \cellcolor{rankthree}0.6225        & 0.7265                             & 0.5963                             & 0.6126                             & 0.6361                             \\
ID        & \textbf{\cellcolor{rankone}0.7391} & \textbf{\cellcolor{rankone}0.7275} & 0.6982                             & 0.6554                             & 0.7990                             & \cellcolor{rankthree}0.7238        & \cellcolor{ranktwo}0.6709          & 0.5872                             & 0.7151                             & \cellcolor{rankthree}0.6050        & 0.6597                             & 0.6476                             \\
FEPoID    & \textbf{\cellcolor{rankone}0.7391} & \textbf{\cellcolor{rankone}0.7275} & \cellcolor{ranktwo}0.7364          & \textbf{\cellcolor{rankone}0.7198} & \cellcolor{ranktwo}0.8483          & \textbf{\cellcolor{rankone}0.7542} & \cellcolor{ranktwo}0.6709          & \textbf{\cellcolor{rankone}0.6514} & \textbf{\cellcolor{rankone}0.7641} & \textbf{\cellcolor{rankone}0.7028} & \cellcolor{ranktwo}0.7413          & \cellcolor{ranktwo}0.7061          \\ \bottomrule
\end{tabular}
}%
\end{table}
% You can have as much text here as you want. The main body must be at most $8$
% pages long. For the final version, one more page can be added. If you want, you
% can use an appendix like this one.

% The $\mathtt{\backslash onecolumn}$ command above can be kept in place if you
% prefer a one-column appendix, or can be removed if you prefer a two-column
% appendix.  Apart from this possible change, the style (font size, spacing,
% margins, page numbering, etc.) should be kept the same as the main body.

\section{Additional Hallucination Detection Results}
\paragraph{Generalization Across Scales and Tuning Strategies}
To evaluate the generalizability of FEPoID beyond instruction-tuned models and standard model scales, we conduct experiments on LLaMA-3.1-8B (base), LLaMA-3.2-3B, and LLaMA-3.2-1B. Results are reported in Tables~\ref{tab:model_8B_base} and~\ref{tab:model_scales}, in which FEPoID consistently selects high-
performing layers and outperforms all baselines.

\begin{figure*}[t]
\centering
\includegraphics[width=0.75\linewidth]{figures/ID_AUROC/without_FST/legend.pdf}
\captionsetup[subfigure]{font=footnotesize, labelfont=footnotesize}
% 控制 tabular 默认列间距（越小越紧）
\setlength{\tabcolsep}{1pt}
\resizebox{\linewidth}{!}{%
  \begin{tabular}{@{}c@{}c@{}c@{}c@{}c@{}c@{}c@{}}
    \raisebox{0pt}[0pt][0pt]{\includegraphics[height=3.7cm]{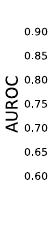}} &
    \subcaptionbox{CoQA}{\includegraphics[height=3.7cm,keepaspectratio]{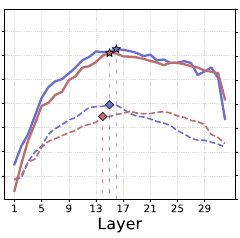}} &
    \subcaptionbox{SQuAD}{\includegraphics[height=3.7cm,keepaspectratio]{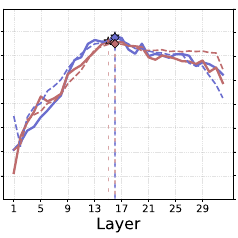}} &
    \subcaptionbox{HotpotQA}{\includegraphics[height=3.7cm,keepaspectratio]{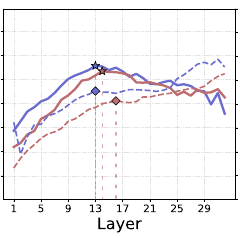}} &
    \subcaptionbox{TriviaQA}{\includegraphics[height=3.7cm,keepaspectratio]{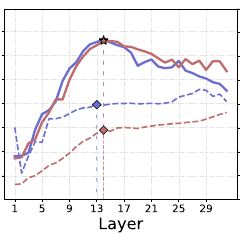}} &
    \subcaptionbox{PsiLoQA}{\includegraphics[height=3.7cm,keepaspectratio]{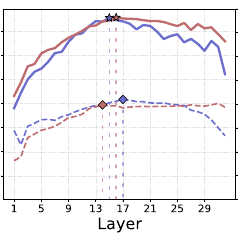}} &
    \raisebox{0pt}[0pt][0pt]{\includegraphics[height=3.7cm]{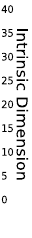}}
  \end{tabular}%d
}

\caption{Layer-wise AUROC and Intrinsic Dimension across QA datasets with FST. Diamond markers indicate the layers selected by FEPoID, and star markers denote the oracle best-performing layers in terms of AUROC. The representations are extracted at the last token of the first generated sentence.}
\label{fig:layer_selection_FST}
\end{figure*}

\paragraph{First-Sentence Truncation Analysis}
To better understand the impact of token-position choice, we report detailed hallucination detection results with first-sentence truncation across datasets in \cref{tab:first_sentence_truncation}. 
Overall, under the FST setting, \ac{our_method} continues to achieve the best average performance across different models.
In addition, \cref{fig:layer_selection_FST} presents layer-wise AUROC curves together with intrinsic dimension estimates when representations are extracted at the last token of the first generated sentence.
Compared to \cref{fig:layer_selection_no_truncation}, the AUROC curves of the two models become noticeably more aligned after applying FST. This observation suggests that different instruction-tuned models tend to capture more consistent and useful information at earlier stages of generation, leading to improved consistency in detection performance.

\begin{figure*}[t]
\centering
\includegraphics[width=0.55\linewidth]{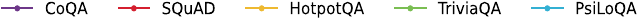}

\captionsetup[subfigure]{font=footnotesize, labelfont=footnotesize}
\setlength{\tabcolsep}{1pt}

\resizebox{\linewidth}{!}{%
  \begin{tabular}{@{}c@{}c@{}c@{}c@{}c@{}c@{}c@{}c@{}}
    \raisebox{0pt}[0pt][0pt]{\includegraphics[height=3.7cm]{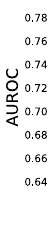}} &
    \subcaptionbox{LlaMA-Instruct (w/o FST)}{\includegraphics[height=3.7cm,keepaspectratio]{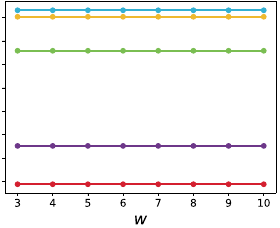}} &
    \raisebox{0pt}[0pt][0pt]{\includegraphics[height=3.7cm]{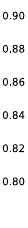}} &
    \subcaptionbox{Mistral-Instruct (w/o FST)}{\includegraphics[height=3.7cm,keepaspectratio]{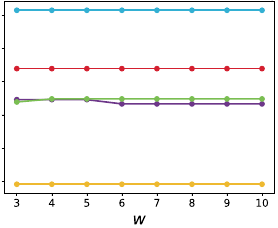}} &
    \raisebox{0pt}[0pt][0pt]{\includegraphics[height=3.7cm]{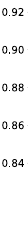}} &
    \subcaptionbox{LlaMA-Instruct (w/ FST)}{\includegraphics[height=3.7cm,keepaspectratio]{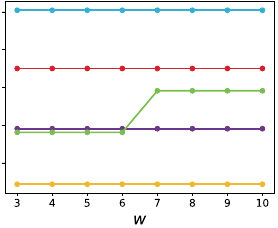}} &
    \raisebox{0pt}[0pt][0pt]{\includegraphics[height=3.7cm]{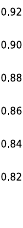}} &
    \subcaptionbox{Mistral-Instruct (w/ FST)}{\includegraphics[height=3.7cm,keepaspectratio]{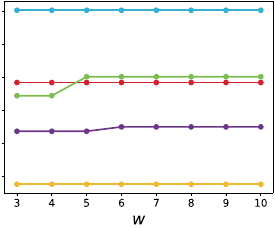}}
  \end{tabular}%
}

\caption{AUROC versus forward horizon $w$ for \ac{our_method} across QA datasets, with and without FST. Results show strong robustness to $w$, with slightly improved performance under FST for larger horizons on some datasets.}
\label{fig:w}
\end{figure*}

\paragraph{Sensitivity to Forward Horizon $w$}
Finally, we examine the sensitivity of FEPoID to the forward horizon parameter $w$ in \cref{fig:w}.
The results demonstrate that FEPoID is robust to a wide range of $w$ values across datasets.
% \paragraph{Model Scale}

\begin{table}[t]
\centering
\caption{Hallucination detection performance (AUROC) across five QA datasets for LLaMA-3.1-8B-Instruct and Mistral-7B-Instruct. For each method, we apply first-sentence truncation for the generated answers. Forward horizon $w$ is set to 7.}
\label{tab:first_sentence_truncation}
\resizebox{\linewidth}{!}{%
\begin{tabular}{cccccccccccccc}
\toprule
\multicolumn{2}{c}{}                                                                        & \multicolumn{6}{c}{LlaMA-3.1-8B-Instruct}                                                                 & \multicolumn{6}{c}{Mistral-7B-Instruct-v0.3}                                                              \\
\multicolumn{2}{c}{}                                                                        & CoQA            & SQuAD           & HotpotQA        & TriviaQA        & PsiLoQA         & Avg             & CoQA            & SQuAD           & HotpotQA        & TriviaQA        & PsiLoQA         & Avg             \\ \midrule
\multicolumn{2}{c}{Pred. Entropy}                                                           & 0.7300          & 0.7850          & 0.7264          & 0.7897          & 0.7880          & 0.7638          & 0.7396          & 0.7633          & 0.7449          & 0.7734          & 0.7574          & 0.7557          \\
\multicolumn{2}{c}{LN-Pred. Entropy}                                                        & 0.6895          & 0.7425          & 0.7427          & 0.7962          & 0.5742          & 0.7090          & 0.6532          & 0.6564          & 0.6970          & 0.7305          & 0.5045          & 0.6483          \\
\multicolumn{2}{c}{Semantic Entropy}                                                        & 0.7015          & 0.7806          & 0.7108          & 0.8148          & 0.7392          & 0.7494          & 0.5772          & 0.6314          & 0.6518          & 0.7603          & 0.6685          & 0.6578          \\
\multicolumn{2}{c}{Lexical Similarity}                                                      & 0.7902          & 0.8433          & 0.7924          & 0.8614          & 0.8061          & 0.8187          & 0.7081          & 0.7270          & 0.7534          & 0.8171          & 0.6907          & 0.7393          \\
\multicolumn{2}{c}{LID}                                                                     & 0.5535          & 0.5731          & 0.5714          & 0.5550          & 0.6648          & 0.5836          & 0.5286          & 0.5784          & 0.5590          & 0.4902          & 0.6600          & 0.5632          \\
\multicolumn{2}{c}{EigenScore}                                                              & 0.7362          & 0.8152          & 0.7271          & 0.8089          & 0.7941          & 0.7763          & 0.7087          & 0.7350          & 0.7357          & 0.7707          & 0.7517          & 0.7404          \\ \midrule
\multirow{7}{*}{\begin{tabular}[c]{@{}c@{}}Hidden-State\\ Probing\end{tabular}} & RankME    & 0.8159          & 0.8106          & \cellcolor{rankthree}{0.8013}          & 0.8228          & 0.8549          & 0.8211          & 0.7746          & 0.7548          & 0.6931          & 0.6763          & 0.8617          & 0.7521          \\
                                                                                & Curvature & \cellcolor{rankthree}{0.8557}          & \cellcolor{rankthree}{0.8625}          & \cellcolor{ranktwo}{ 0.8119}    & \cellcolor{rankthree}{0.8287}          & \cellcolor{rankthree}{0.9160}          & \cellcolor{rankthree}{0.8550}          & \cellcolor{ranktwo}{ 0.8474}    & \cellcolor{rankthree}{0.8704}          & \cellcolor{ranktwo}{ 0.8088}    & \cellcolor{ranktwo}{ 0.8680}    & \cellcolor{ranktwo}{ 0.9264}    & \cellcolor{rankthree}{0.8642}          \\
                                                                                & Val Loss  & \cellcolor{rankone}\textbf{0.8621} & \cellcolor{ranktwo}{ 0.8865}    & \cellcolor{rankone}\textbf{0.8287} & \cellcolor{ranktwo}{ 0.8731}    & \cellcolor{rankone}\textbf{0.9238} & \cellcolor{ranktwo}{ 0.8748}    & \cellcolor{rankthree}{0.8430}          & \cellcolor{rankone}\textbf{0.8784} & \cellcolor{rankthree}{0.8087}          & \cellcolor{rankone}\textbf{0.8805} & \cellcolor{rankone}\textbf{0.9216} & \cellcolor{ranktwo}{ 0.8664}    \\
                                                                                & RGN       & 0.7167          & 0.8104          & 0.7292          & 0.7772          & 0.8112          & 0.7689          & 0.8346          & 0.7929          & 0.7631          & \cellcolor{rankthree}{0.8176}          & 0.8797          & 0.8176          \\
                                                                                & SNR       & 0.7602          & 0.6693          & 0.7723          & 0.6408          & 0.8001          & 0.7285          & 0.7453          & 0.7112          & 0.7259          & 0.7115          & 0.8274          & 0.7443          \\
                                                                                &ID    & \cellcolor{ranktwo}{ 0.8581}    & \cellcolor{rankone}\textbf{0.8900} & 0.7723          & 0.8066          & \cellcolor{ranktwo}{ 0.9208}    & 0.8496          & \cellcolor{ranktwo}{ 0.8474}    & \cellcolor{ranktwo}{ 0.8769}    & 0.7631          & \cellcolor{rankthree}{0.8176}          & 0.8942          & 0.8398          \\
                                                                                & FEPoID  & \cellcolor{ranktwo}{ 0.8581}    & \cellcolor{rankone}\textbf{0.8900} & \cellcolor{rankone}\textbf{0.8287} & \cellcolor{rankone}{\textbf{0.8782}} & \cellcolor{ranktwo}{ 0.9208}    & \cellcolor{rankone}\textbf{0.8752} & \cellcolor{rankone}\textbf{0.8501} & \cellcolor{ranktwo}{ 0.8769}    & \cellcolor{rankone}\textbf{0.8153} & \cellcolor{rankone}\textbf{0.8805} & \cellcolor{rankthree}{0.9207}          & \cellcolor{rankone}\textbf{0.8687} \\ \midrule
\end{tabular}}%
\end{table}

% \begin{figure}[t]
%     \centering
%     \includegraphics[width=\textwidth]{figures/pdfs/windowsize.pdf}
%     \caption{AUROC versus forward horizon $w$ for \ac{our_method} across QA datasets, with and without FST. Results show strong robustness to $w$, with slightly improved performance under FST for larger horizons on some datasets.}
%     \label{fig:w}
% \end{figure}
\section{Linear Probe}
\label{linear_probe}
To support that the FEPoID-selected layer encodes abstract semantic information, we additionally run two sets of linear probing experiments using logistic regression on the frozen hidden states, comparing the FEPoID-selected layer (index 0) against its neighboring layers (index $\pm1$, $\pm2$, $\pm3$). 
% We provide the results below which further validate our statements.
\paragraph{Factual Correctness Probing}
We train a logistic regression probe at each neighboring layer to predict whether the model's generated answer is factually correct, using the binary correctness labels already available in our experimental pipeline. We report AUROC across all five QA datasets for LLaMA-3.1-8B in \cref{tab:linear_probing}. 
To further assess whether the FEPoID-selected layer encodes abstract semantic categories beyond task-specific signals, we consider three external benchmarks:
\begin{enumerate}
    \item Odd Man Out~\citep{conneau-etal-2018-cram}: A binary semantic coherence task where each sentence is labeled as either Original (well-formed) or Changed (a noun or verb has been replaced by a random word of the same part of speech). Correctly distinguishing O from C requires genuine semantic world knowledge rather than surface-level features.
    \item AG’s News~\citep{zhang2015character}: A 4-class topic classification task over news articles (World, Sports, Business, Science and Technology), probing whether the layer encodes semantic topic categories.
    \item DBPedia~\citep{zhang2015character}: A 14-class ontology classification task over Wikipedia entity descriptions, requiring the probe to distinguish fine-grained semantic categories such as Artist, Athlete, Animal, Building, and Film, categories that share surface-level features but differ in semantic type.
\end{enumerate}
For all three tasks, we feed each sentence directly into the frozen LLaMA-3.1-8B and conduct probing using logistic regression. 

We report the AUROC results for both sets of experiments in \cref{tab:linear_probing}. The FEPoID-selected layer (index 0) consistently achieves the highest or near-highest AUROC across datasets, while performance consistently degrades as we move to deeper or shallower neighboring layers. 
Both the factual correctness probing and semantic category probing results confirm that the FEPoID-selected layer is not merely a coincidental choice for hallucination detection, but is the layer where abstract semantic information is encoded. We hope you can take our response into account and consider raising your score in the final assessment.
\begin{table}[ht]
\centering
\caption{Linear probing accuracy across layers. The FEPoID-selected layer (index 0) is compared against neighboring layers. Best results per column are in bold.}
\label{tab:linear_probing}
\footnotesize
\begin{tabular}{ccccccccc}
\toprule
Layer & CoQA & SQuAD & HotpotQA & TriviaQA & PsiLoQA & Odd Man Out & AG's News & DBPedia \\
\midrule
$-3$ & 0.7505 & 0.8250 & 0.7635 & 0.8494 & 0.8293 & 0.8180 & 0.9830 & 0.8318 \\
$-2$ & 0.7870 & 0.7824 & 0.7722 & 0.8543 & 0.8157 & 0.8207 & 0.9836 & 0.8227 \\
$-1$ & 0.8017 & 0.8308 & 0.7604 & 0.8536 & 0.8692 & 0.8217 & 0.9835 & \textbf{0.8398} \\
0 (FEPoID) & 0.8109 & \textbf{0.8582} & \textbf{0.7791} & 0.8555 & \textbf{0.8849} & \textbf{0.8252} & \textbf{0.9837} & 0.8386 \\
1 & \textbf{0.8130} & 0.8313 & 0.7761 & \textbf{0.8560} & 0.7920 & 0.8226 & 0.9833 & 0.8175 \\
2 & 0.8066 & 0.8419 & 0.7722 & 0.8533 & 0.8555 & 0.8185 & \textbf{0.9837} & 0.7969 \\
3 & 0.7832 & 0.8566 & 0.7568 & 0.8466 & 0.8726 & 0.8170 & 0.9828 & 0.7748 \\
\bottomrule
\end{tabular}

\end{table}
\section{Generalization to Vision Tasks}

\begin{table}[t]
\centering
\caption{Results on the \textbf{LlaMA-3.1-8B base} model with $w=7$. The representations are extracted with FST. FEPoID maintains strong performance on a non-instruction-tuned 
model, demonstrating that our method \emph{generalizes} across both base and 
instruction-tuned models. }
\label{tab:model_8B_base}
\begin{tabular}{ccccccc}
\toprule
          & CoQA                               & SQuAD                              & HotpotQA                           & TriviaQA                           & PsiloQA                            & Avg                                \\ \midrule
RankME    & 0.7624                             & 0.8540                             & 0.6856                             & 0.6612                             & 0.8398                             & 0.7606                             \\
Curvature & \cellcolor{rankthree}0.7951        & \cellcolor{rankthree}0.8613        & \cellcolor{ranktwo}0.7839          & \cellcolor{rankone}0.8596          & \cellcolor{ranktwo}0.9049          & \cellcolor{rankthree}0.8410        \\
Val Loss  & \cellcolor{ranktwo}0.8162          & \cellcolor{ranktwo}0.8870          & \cellcolor{rankthree}0.7735        & \textbf{\cellcolor{rankone}0.8799} & \textbf{\cellcolor{rankone}0.9051} & \cellcolor{ranktwo}0.8523          \\
RGN       & 0.7285                             & 0.7305                             & 0.7499                             & \cellcolor{rankthree}0.7684        & 0.8583                             & 0.7671                             \\
SNR       & 0.6565                             & 0.7230                             & 0.6552                             & 0.6370                             & 0.7999                             & 0.6943                             \\
ID        & \textbf{\cellcolor{rankone}0.8264} & \textbf{\cellcolor{rankone}0.8884} & 0.7232                             & \cellcolor{rankthree}0.7684        & \cellcolor{rankthree}0.8985        & 0.8210                             \\
FEPoID    & \textbf{\cellcolor{rankone}0.8264} & \textbf{\cellcolor{rankone}0.8884} & \textbf{\cellcolor{rankone}0.7972} & \textbf{\cellcolor{rankone}0.8799} & \cellcolor{rankthree}0.8985        & \textbf{\cellcolor{rankone}0.8581} \\ \bottomrule
\end{tabular}
\end{table}

\begin{table}[t]
\centering
\caption{Results on different model scales, where representations are extracted with FST. Forward horizon is set to $w=7$ for \textbf{LlaMA-3.2-3B} and $w = 3$ for \textbf{LlaMA-3.2-1B}. Top-3 results are highlighted, with darker color indicating better performance. FEPoID achieves consistently strong performance across both model scales, demonstrating its \emph{generalizability} to models of varying scales. 
}
\label{tab:model_scales}
\footnotesize
\resizebox{\textwidth}{!}{%
\begin{tabular}{ccccccccccccc}
\toprule
          & \multicolumn{6}{c}{LlaMA-3.2-3B}                                                                                                                                                                                   & \multicolumn{6}{c}{LlaMA-3.2-1B}                                                                                                                                                                                   \\
          & CoQA                               & SQuAD                              & HotpotQA                           & TriviaQA                           & PsiloQA                            & Avg                                & CoQA                               & SQuAD                              & HotpotQA                           & TriviaQA                           & PsiloQA                            & Avg                                \\ \midrule
RankME    & 0.6265                             & 0.7338                             & 0.7083                             & 0.6798                             & 0.7950                             & 0.7087                             & 0.5813                             & 0.6288                             & 0.7009                             & 0.6547                             & 0.7151                             & 0.6561                             \\
Curvature & \cellcolor{ranktwo}0.7737          & \cellcolor{rankthree}0.8163        & \cellcolor{rankthree}0.7779        & 0.8200                             & \cellcolor{rankthree}0.8687        & \cellcolor{rankthree}0.8113        & \cellcolor{rankthree}0.7190        & \cellcolor{ranktwo}0.7340          & \cellcolor{ranktwo}0.7957          & \cellcolor{ranktwo}0.7876          & \cellcolor{ranktwo}0.8097          & \cellcolor{rankthree}0.7692        \\
Val Loss  & \textbf{\cellcolor{rankone}0.7919} & \textbf{\cellcolor{rankone}0.8534} & \cellcolor{ranktwo}0.7900          & \cellcolor{ranktwo}0.8441          & \cellcolor{ranktwo}0.8822          & \cellcolor{ranktwo}0.8323          & \cellcolor{ranktwo}0.7329          & \textbf{\cellcolor{rankone}0.7566} & \textbf{\cellcolor{rankone}0.8030} & \cellcolor{rankthree}0.7750        & \textbf{\cellcolor{rankone}0.8323} & \cellcolor{ranktwo}0.7800          \\
RGN       & 0.6786                             & 0.6333                             & 0.7493                             & 0.7027                             & 0.8295                             & 0.7187                             & 0.6421                             & 0.6499                             & 0.7485                             & 0.7199                             & 0.7788                             & 0.7078                             \\
SNR       & \cellcolor{rankthree}0.7639        & 0.6333                             & \cellcolor{rankthree}0.7779        & \cellcolor{rankthree}0.8337        & 0.7665                             & 0.7550                             & 0.6859                             & 0.5973                             & \cellcolor{rankthree}0.7694        & 0.6547                             & \cellcolor{ranktwo}0.8097          & 0.7034                             \\
ID        & \textbf{\cellcolor{rankone}0.7919} & \cellcolor{ranktwo}0.8461          & 0.7309                             & 0.7027                             & \textbf{\cellcolor{rankone}0.8843} & 0.7912                             & \textbf{\cellcolor{rankone}0.7356} & \cellcolor{rankthree}0.7074        & 0.7485                             & 0.7199                             & \cellcolor{rankthree}0.7897        & 0.7402                             \\
FEPoID    & \textbf{\cellcolor{rankone}0.7919} & \cellcolor{ranktwo}0.8461          & \textbf{\cellcolor{rankone}0.7992} & \textbf{\cellcolor{rankone}0.8479} & \textbf{\cellcolor{rankone}0.8843} & \textbf{\cellcolor{rankone}0.8339} & \textbf{\cellcolor{rankone}0.7356} & \textbf{\cellcolor{rankone}0.7566} & \textbf{\cellcolor{rankone}0.8030} & \textbf{\cellcolor{rankone}0.8025} & \textbf{\cellcolor{rankone}0.8323} & \textbf{\cellcolor{rankone}0.7860} \\ \bottomrule
\end{tabular}
}%
\end{table}
We evaluate \ac{our_method} on CIFAR-10 using an ImageNet-pretrained ViT. For each layer, we use the \texttt{[CLS]} token representation to train and evaluate the downstream MLP.

\begin{figure}[t]
    \centering
    \includegraphics[width=0.5\linewidth]{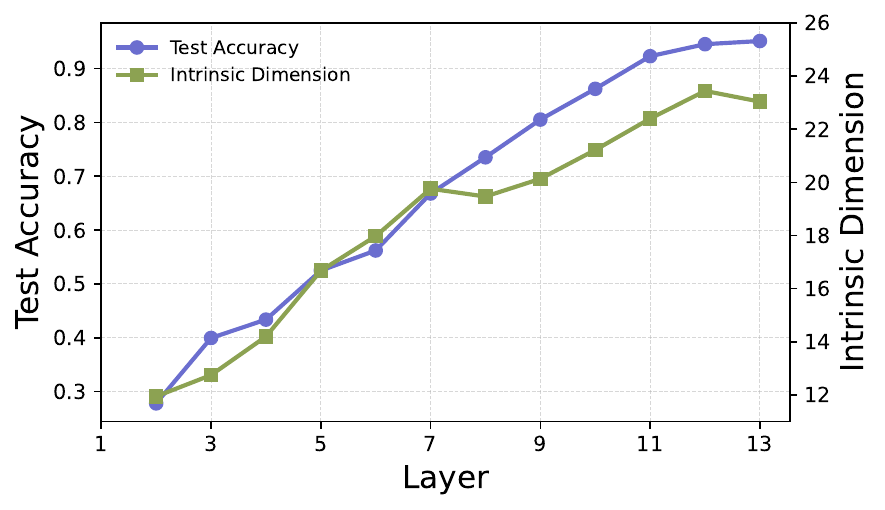}
    \caption{Layer-wise accuracy and intrinsic dimension for image classification. \ac{our_method} picks the last second layer with $w\in \{2,3,4,5\}$.}
    \label{fig:image_task}
\end{figure}

% Following the hidden-state probing paradigm, the ViT backbone is kept frozen, and downstream classifiers are trained on representations extracted from individual transformer blocks.
As shown in \cref{fig:image_task}, test accuracy increases monotonically with network depth and reaches its maximum at the final transformer block.
This behavior contrasts with hallucination detection, where intermediate layers often yield the strongest probe performance (\cref{fig:motivation,fig:layer_selection_no_truncation}).

Consistent with this trend, the estimated intrinsic dimension increases steadily across layers and peaks at the penultimate layer.
Importantly, with $w\in[2,5]$, \ac{our_method} accurately selects this layer, closely matching the oracle best-performing layer.
These results demonstrate that \ac{our_method} reliably identifies informative layers for vision tasks, extending its applicability beyond language-based settings.
%%%%%%%%%%%%%%%%%%%%%%%%%%%%%%%%%%%%%%%%%%%%%%%%%%%%%%%%%%%%%%%%%%%%%%%%%%%%%%%
%%%%%%%%%%%%%%%%%%%%%%%%%%%%%%%%%%%%%%%%%%%%%%%%%%%%%%%%%%%%%%%%%%%%%%%%%%%%%%%

\end{document}